\documentclass[lettersize,journal]{IEEEtran}
\usepackage{amsmath,amsfonts}
\usepackage[inkscapelatex=false]{svg}
\usepackage{colortbl}
\usepackage{algorithmic}
\usepackage{algorithm}
\usepackage{lineno}
\usepackage[caption=false,font=normalsize,labelfont=sf,textfont=sf]{subfig}
\usepackage{textcomp}
\usepackage{stfloats}
\usepackage{url}
\usepackage{verbatim}
\usepackage{graphicx}
\usepackage{cite}
\usepackage{multirow}
\usepackage{units}
\usepackage[caption=false,font=normalsize,labelfont=sf,textfont=sf]{subfig}
\makeatletter

\newcommand{\Rmnum}[1]{\expandafter\@slowromancap\romannumeral #1@}
\makeatother
\begin{document}
\title{Efficient Large-Scale Urban Parking Prediction: Graph Coarsening Based on Real-Time Parking Service Capability}

\author{Yixuan Wang$^{\&1,2}$, Zhenwu Chen$^{\&3}$, Kangshuai Zhang$^{1}$, Yunduan Cui$^{1}$, \textit{Member}, \textit{IEEE}, Yang Yang$^{4,5,6}$, \textit{Fellow}, \textit{IEEE}, and Lei Peng$^{*1}$, \textit{Member}, \textit{IEEE}
% \author{Yixuan Wang$^{\&}$, Zhenwu Chen$^{\&}$, Kangshuai Zhang, Yunduan Cui, \textit{Member}, \textit{IEEE}, and Lei Peng$^{*}$, \textit{Member}, \textit{IEEE}

        % <-this % stops a space
\thanks{$^{*}$Corresponding author: Lei Peng (email: lei.peng@siat.ac.cn)}
\thanks{$^{\&}$Yixuan Wang and Zhenwu Chen are co-first authors}
\thanks{$^{1}$Shenzhen Institutes of Advanced Technology, Chinese Academy of Sciences, Shenzhen 518055, China}
\thanks{$^{2}$University of Chinese Academy of Sciences, Beijing 100049, China}
\thanks{$^{3}$Shenzhen Urban Transport Planning Center Co.,Ltd. shenzhen 518057, China}
\thanks{$^{4}$IoT Thrust and the Research Center for Digital World with Intelligent Things (DOIT) at HKUST, Guangzhou 511453, China}
\thanks{$^{5}$Peng Cheng Laboratory, Shenzhen 518055, China}
\thanks{$^{6}$Terminus Group, Beijing 100027, China}
% \thanks{Yixuan Wang is with Shenzhen Institute of Advanced Technology, Chinese Academy of Sciences, Shenzhen 518055, China, also with University of Chinese Academy of Sciences, Beijing 100049, China (email: yx.wang5@siat.ac.cn)}
% \thanks{Zhenwu Chen is with Shenzhen Urban Transport Planning Center Co.,Ltd. shenzhen 518057, China (email: czw@sutpc.com)}
% \thanks{Kangshuai Zhang, Yunduan Cui and Lei Peng are with Shenzhen Institutes of Advanced Technology, Chinese Academy of Sciences, Shenzhen 518055, China (email: ks.zhang@siat.ac.cn; yd.cui@siat.ac.cn; lei.peng@siat.ac.cn)}
% \thanks{Yang Yang is with the IoT Thrust and the Research Center for Digital World with Intelligent Things (DOIT) at HKUST (Guangzhou), China, also with Peng Cheng Laboratory, Shenzhen 518055, China, and also  with Terminus Group, Beijing 100027, China (email: yyiot@hkust-gz.edu.cn)}
% \thanks{$^{\&}$Yixuan Wang and xxx are co-first authors}
% \thanks{$^{1}$University of Chinese Academy of Sciences. Beijing 100049, China}%
% \thanks{$^{2}$Shenzhen Institute of Advanced Technology, Chinese Academy of Sciences. Shenzhen 518055, China}
% \thanks{$^{4}$The IoT Thrust and the Research Center for Digital World with Intelligent Things (DOIT) at HKUST (Guangzhou), China}
% \thanks{$^{5}$Peng Cheng Laboratory. Shenzhen 518055, China}
% \thanks{$^{6}$Peng Cheng Laboratory. Shenzhen 518000, China}
}

\markboth{Journal of \LaTeX\ Class Files,~Vol.~14, No.~8, August~2021}%
{Shell \MakeLowercase{\textit{et al.}}: A Sample Article Using IEEEtran.cls for IEEE Journals}
\maketitle

\begin{abstract}
With the sharp increase in the number of vehicles, the issue of parking difficulties has emerged as an urgent challenge that many cities need to address promptly. In the task of predicting large-scale urban parking data, existing research often lacks effective deep learning models and strategies. To tackle this challenge, this paper proposes an innovative framework for predicting large-scale urban parking graphs leveraging real-time service capabilities, aimed at improving the accuracy and efficiency of parking predictions. Specifically, we introduce a graph attention mechanism that assesses the real-time service capabilities of parking lots to construct a dynamic parking graph that accurately reflects real preferences in parking behavior. To effectively handle large-scale parking data, this study combines graph coarsening techniques with temporal convolutional autoencoders to achieve unified dimension reduction of the complex urban parking graph structure and features. Subsequently, we use a spatiotemporal graph convolutional model to make predictions based on the coarsened graph, and a pre-trained autoencoder-decoder module restores the predicted results to their original data dimensions, completing the task. Our methodology has been rigorously tested on a real dataset from parking lots in Shenzhen. The experimental results indicate that compared to traditional parking prediction models, our framework achieves improvements of 46.8\% and 30.5\% in accuracy and efficiency, respectively. Remarkably, with the expansion of the graph's scale, our framework's advantages become even more apparent, showcasing its substantial potential for solving complex urban parking dilemmas in practical scenarios.
\end{abstract}


\section{INTRODUCTION}
\IEEEPARstart{U}{}rban transportation systems are facing serious challenges due to the continuous increase in the number of vehicles. Recent statistics\cite{1} reveal that vehicle ownership in China escalated to 417 million in 2022, complemented by a striking addition of 34.78 million new registrations within the year. Particularly at the urban scale, as many as 13 cities saw an increase in car ownership by more than 3 million vehicles, including metropolises like Beijing, Shanghai, and Shenzhen. This surge in vehicle numbers not only leads to traffic congestion but also severely exacerbates parking issues. The expansion of urban parking facilities and spaces cannot keep pace with the growth in vehicle numbers, making it exceedingly difficult to find parking in city centers and commercial areas.

To alleviate the parking difficulty issue, urban-level smart parking platforms \cite{2} have emerged. By leveraging cutting-edge technologies such as digital twins, the Internet of Things, and cloud computing \cite{3}, these platforms have successfully integrated urban parking facility resources, effectively solving the problem of information asymmetry between users and available parking spaces. Taking the Wuhan smart parking project \cite{4} as an example, by the end of 2022, the platform had integrated more than 4,000 parking lots, including approximately 480,000 parking space data. With its portability, the platform offers convenience to citizens, who can easily check real-time parking availability in different areas through their mobile phones and use services such as parking reservation \cite{5} and parking navigation \cite{6},\cite{7},\cite{8} to achieve precise parking. In these platforms, the key to providing smart parking services lies in parking prediction technology \cite{9}. Indeed, parking prediction has always been a hot research topic in urban parking technology \cite{10}. By effectively predicting the availability of parking spaces in future time periods, it can not only enhance user experience but also improve parking turnover, optimize drivers' parking behavior, and thereby help to allocate urban parking resources rationally.

Recent studies have framed such parking prediction tasks as spatiotemporal graph modeling problems \cite{11}. On one hand, this is because parking lots are connected through urban road networks, forming unique parking graphs \cite{12}, for which graph convolutional networks can be utilized to capture the spatial characteristics of the parking graph \cite{13}. On the other hand, given the tidal characteristics of parking flow, recurrent neural networks can be employed to fit the time series regression of parking data \cite{14}. This approach, which considers both spatial and temporal dependencies, is known as spatiotemporal Graph Convolutional Networks (GCNs).

However, for the vehicle-dense cities mentioned above, with the rapid development of wireless sensors and the widespread use of vehicles \cite{15}, a massive influx of real-time parking data continuously flows into urban-level parking platforms. This leads to a dramatic increase in the volume of parking data managed  by existing platforms and the coverage range of the parking Internet of Things network \cite{16}. The rapid influx of such information necessitates an unsustainable amount of time to train spatiotemporal GCNs at the required scale for large-scale urban parking prediction.

To reduce the resources and time costs involved in training the aforementioned models, scholars have been dedicated to finding efficient graph dimensionality reduction techniques \cite{17}. Among various studies, graph coarsening \cite{18} stands out as one of the mainstream methods that can achieve this aim. Its core concept involves merging certain nodes within the graph into a hypernode and then using these hypernodes to construct a coarsened graph, thereby enhancing the learning efficiency and scalability of large-scale graphs \cite{19}. However, a prevailing issue is that existing coarsening methods struggle to be directly applied to complex parking scenarios:
 % \vspace{-0.1mm}
\begin{enumerate}
\item The quality of the input graph significantly determines the performance of downstream tasks\cite{20}. For parking graphs, a high-quality parking graph is not only an abstraction of spatial topology but also a adequate representation of parking behavior and decision-making. Neglecting either aspect will limit the understanding of parking scenarios, thereby negatively affecting the effectiveness of subsequent coarsening tasks as well as the accuracy and robustness of prediction tasks.
\item Most of the existing coarsening methods only consider reducing the size of the graph structure \cite{21}, and the node features of the coarsened graph still being obtained by concatenating the original data. This means that the amount of data has not changed. If the model is trained directly based on the coarsened parking  graph, the training overhead is not significantly reduced. Moreover, since the coarsened parking graph loses the original topology, the merged data corresponding to the hypernodes also lose their spatial features, leading to a decrease in prediction accuracy.
\end{enumerate}

Graph attention networks (GAT) have attracted much attention in the transportation field \cite{22} over recent years. By selectively focusing on specific nodes and discerning the importance among various nodes, these mechanisms help models to capture deeper hidden information in the traffic scenarios \cite{23}, thereby elevating predictive accuracy. The ParkingRank algorithm \cite{24}, which assesses the real-time service capability of parking lots using their static attributes, has been widely applied in parking recommendations \cite{25}, parking guidance \cite{2}, and other related tasks. We posit that this methodology can be interpreted as a preference-based attention mechanism, adept at capturing the service capability variations across different parking lots by evaluating their significance. This nuanced attention mechanism, by thoroughly considering node characteristics alongside topological connections, offers enriched guidance for the coarsening process, enhancing the model's overall utility and effectiveness.

Meanwhile, autoencoders (AEs) have been widely used in traffic data compression and reconstruction tasks due to their theoretical ability to achieve lossless compression \cite{26}. Specifically, the encoding module is able to learn the high-dimensional sparse information in traffic data and successfully encode it into a low-dimensional dense tensor \cite{27} for more efficient traffic prediction, while the decoding module is tasked with reconstructing the condensed representations back to their original data form \cite{28}. In this framework, we argue that the introduction of an AE can compensate for the lack of compression of node features in the graph coarsening process. Through the synergy of the encoding and decoding modules, a unified dimensionality reduction of the parking graph structure and features can be achieved, thus providing a more efficient and accurate data representation for the downstream parking prediction task. In summary, the main contributions of this paper can be summarized as follows:
 % \vspace{-0.3mm}
\begin{itemize}
\item We have proposed a method for constructing parking graphs that leverages a ParkingRank-based attention mechanism. Specifically, this method integrates real-time assessments of parking lot service capacities into a graph attention network, generating a parking graph that accurately reflects real-world parking behavior preferences. This graph not only captures the complex characteristics of parking scenarios but also enhances the interpretability of the graph attention mechanism in analyzing intricate parking behavior patterns. Additionally, it adapts seamlessly to dynamic changes within parking environments, thereby providing a solid foundation for downstream parking graph coarsening as well as prediction tasks.
\item We introduce a novel framework for parking prediction that employs a graph coarsening techniques and temporal convolutional autoencoder\cite{29}, designed to diminish the resource and time expenditures associated with urban parking prediction models. The scheme can make up for the shortcomings of traditional coarsening methods that neglect the dimensionality reduction of node features, and realize the unified dimensionality reduction of urban parking graph structure and features. Moreover, by incorporating temporal convolutional networks in the encoding-decoding phase, our approach not only significantly enhances the dimensionality reduction and reconstruction capabilities for parking time series data but also ensures the accuracy of the parking prediction task. Additionally, the compact data volume within each hypernode allows for the parallel processing of encoding-decoding operations across different sets of parking time series data, further boosting the overall training efficiency of the parking prediction task.
\end{itemize}

This paper is structured as follows: Section II provides a concise overview of spatiotemporal GCNs for parking prediction, techniques for dimensionality reduction in large-scale graphs, and the applications of Autoencoders (AE). Section III delves into the specifics of our innovative ParkingRank graph attention model and outlines our parking prediction methodology, which leverages TCN-AE for coarsening. Section IV details the experimental framework and analysis employed to assess the effectiveness of our proposed approach. We conclude by summarizing the key insights and contributions of this study.

\section{RELATED WORK}
\subsection{spatiotemporal Graph Convolutional Models}
spatiotemporal GCNs \cite{30} have emerged as a powerful and prevalent approach within the realm of traffic prediction, adept at navigating the intricacies of traffic graphs by integrating both spatial and temporal dependencies. Unlike relying solely on GCNs or RNNs, spatiotemporal GCNs excel at dissecting time-series data embedded in graph structures \cite{31}, thereby yielding more precise prediction. For instance, Zhao et al. \cite{32} introduced the T-GCN model, ingeniously combining GCN with GRU to understand the intricate topology of traffic networks and the dynamic shifts in traffic data, enhancing the capture of spatiotemporal characteristics more effectively. In a similar vein, Sun et al. \cite{33} unveiled DDSTGCN, leveraging TCN to track variations in traffic states alongside GCN for topology learning, markedly boosting traffic prediction accuracy. In recent years, scholars have suggested employing attention mechanisms to bolster the extraction of spatiotemporal features, thereby extending the model's interpretability. Furthermore, Li et al. \cite{34} developed AST-GAT, incorporating multi-head graph attention and attention-based LSTM modules to intricately map the spatiotemporal interdependencies among road segments, aiming to refine the precision of traffic prediction.

Due to the inherent structural and functional similarities between urban transportation networks and urban parking networks, both involve complex spatial relationships and dynamically changing temporal characteristics. Leveraging advanced spatiotemporal graph convolutional models from the field of traffic prediction, we believe can effectively enhance the accuracy of parking prediction. However, in the domain of parking prediction, the industrial sector continues to prioritize concerns regarding time costs and resource consumption, with a current lack of effective solutions. These challenges will limit the feasibility and effectiveness of the aforementioned spatiotemporal graph convolutional models in practical industrial applications.

 % \vspace{-3mm}
\subsection{Techniques for Dimensionality Reduction in Large-Scale Graphs}
As the scale of graphs continues to expand, finding a universal method that can simplify the structure while preserving key attributes becomes increasingly important. Simplified graph representations not only facilitate storage but also offer efficiency in approximate algorithms \cite{19}. There are mainly two methods for simplifying graphs: The first method, known as graph sparsification \cite{35}, approximates the maintenance of distance relationships between node pairs by removing edges from the graph. However, for highly structured parking graphs, this method might lose the original graph's connectivity. An improved method is the K-neighbors Sparsifier \cite{36}, which performs local sparsification on each node by setting a threshold, but still introduces high computational complexity. With the growing popularity of deep learning, Wu et al. \cite{37} proposed GSGAN, which generates new graphs through GANs to preserve the community structure of the original graph, but it introduces edges that do not exist in the original graph and contradicts reality.

The second approach is known as graph coarsening \cite{38}, which aims to reduce the number of nodes to a subset of the original nodes while maintaining the original graph properties. The key to graph coarsening \cite{20} lies in the ability to accurately measure the variations before and after coarsening. The GraClus \cite{39} technique, and the MCCA \cite{40} algorithm all employ the idea of greediness by constantly compacting the connected regions of the original graph. However, these approaches lack guarantees for global optimization. Recently, spectral graph theory \cite{21} proved that graphs with similar spectra are usually considered to have similar topologies. Therefore, the idea of minimizing the spectral distance can be introduced into the coarsening process, using the eigenvalues of the graphs to measure the structural similarity and ensuring that the generated coarsened graphs can maintain the properties of the original graphs.

However, we have observed that current graph coarsening methods primarily focus on reducing the complexity of the graph structure while often neglecting the dimensionality of node features. This approach does not effectively reduce the actual volume of parking data in large-scale urban parking prediction tasks. Although the parking network structure has been somewhat simplified through graph coarsening techniques, the training overhead for downstream parking prediction models has not decreased as significantly as anticipated.
 % \vspace{-3mm}
\subsection{Applications of Autoencoders}
% \vspace{1mm}
AEs have demonstrated remarkable proficiency in managing intricate spatiotemporal data relevant to transportation systems \cite{41}. Typically employing symmetrically structured encoders and decoders for unsupervised learning, these AEs are adept not only at distilling input data into compact representations of potential vectors \cite{26}, but also at restoring the data to its original dimensions. They achieve this by optimizing the reconstruction of the data's objective, thereby minimizing loss \cite{28}. For instance, the TGAE introduced by Wang et al. \cite{41} adeptly captures the underlying patterns of traffic flow through a spatiotemporal AE, facilitating the prediction of future traffic information. Similarly, the TCN-AE devised by Mo et al. \cite{42} leverages TCNs within its encoding module to extract features from temporal data, subsequently utilizing the decoding module to map latent representations back to the original data space, thus enabling anomaly detection.

Inspired by the capabilities of  AEs in dimensionality reduction and key feature extraction of spatiotemporal data, we propose the integration of  AEs into large-scale urban parking prediction tasks, aiming to address the limitations of traditional coarsening methods in handling parking lot node features. Additionally, by leveraging the advantages of TCN in processing complex temporal data, AEs can more accurately capture and analyze the parking patterns within different parking lots, thereby significantly enhancing the performance and depth of understanding of downstream parking prediction models.

\section{METHODOLOGY}
\subsection{Problem Definition}
In this paper, our goal is to effectively reduce the time and resource costs of parking prediction models by achieving a unified reduction in the dimensions of the structure and feature data of urban parking graphs, while ensuring the accuracy of parking prediction tasks. The urban parking graph can be represented as $G=(V,E,A)$, where $V$ denotes the set of parking nodes, $E$ denotes the set of edges, and $A \in \mathbb{R}^{N \times N}$ denotes the directed adjacency matrix, $N=|V|$.

The parking prediction problem \cite{11} can be interpreted as: given a parking graph $G$ and historical parking data $X_{(t-T+1):t}$ for $T$ time slices before time $t$, learning a function $F$ to predict future parking data $\hat{Y}_{(t+1):(t+T')}$ for $T'$ time slices after time $t$, as shown in Equation 1.
\begin{equation}
    \hat{Y}_{(t+1):(t+T')} = F(X_{(t-T+1):t}, G),
    \label{eq1}
\end{equation}
where the parking data $X=\{x_1, x_2, \ldots, x_T\} \in \mathbb{R}^{T \times N \times F}$ represents the parking flow information observed at $T$ time slices for all parking lots. Each $x_t$ corresponds to a time slice and includes $N$ parking lot nodes, each with $F$ dimensional features, including the longitude, latitude, openness to the public, charging situation, and real-time occupancy rate of the parking lot.

\IEEEpubidadjcol

\subsection{Parking Service Capacity Assessment}
Given the complexity of the parking data $X$ directly inputting the high-dimensional data $x_t$ into GAT would lead to considerable computational redundancy, thereby increasing the model's time cost. To address this, we adopt a dimensionality reduction strategy that involves extracting and abstracting certain features of the parking lots. This effectively reduces the computational burden and enhances the model's computational efficiency.

The ParkingRank algorithm \cite{24} is an algorithm to quantitatively assess the service capability of different parking lots. By integrating complex parking information such as the total number of parking spaces in the parking lot, the degree of openness to the public, and the price of parking, it can help drivers to understand the actual situation of the parking lot in a more comprehensive way, so that they can make informed choices in the parking decision-making process. Specifically, the algorithm primarily focuses on three aspects to describe the real-time service capacity of parking lots, as shown in Equation 2: 
\begin{itemize}
    \item Parking lot service range: This aspect considers which types of vehicles are allowed to park in the parking lot. For example, parking lots at shopping centers may be open to all vehicles, while those in residential areas may only serve residents. Therefore, parking lots with a broader service scope generally have stronger service capabilities.
    \item Total number of parking spaces: The more internal parking spaces a parking lot has, the stronger its service capacity usually is.
    \item Price of parking: Higher parking prices may reduce the number of vehicles able to afford parking fees. Thus, expensive prices may lower the service capacity of the parking lot. 
\end{itemize}

% \vspace{-5mm}
\begin{equation}
PR_i = \exp(s_i)\frac{ \left(\nicefrac{y_i}{\|y\|} \right)}{1 + \nicefrac{z_i}{\|z\|}},
\label{eq2}
\end{equation}
where $PR_i$ represents the service capacity of each parking lot; $s_i \in [0,1]$ denotes the service range, which is used to measure the degree of openness of the parking lot, from the public parking lot ($s_i = 1$) to the private parking lot ($s_i=0$); $y_i$ and  $z_i$ represent the total number of parking spaces and the price of parking, respectively. $\|y\|$ and $\|z\|$ are the first-order paradigms for the  $y$- and $z$-column vectors. 

We believe that the process of quantifying the service capability of a parking lot is essentially a process of extracting the redundant parts of the parking lot features $x_t$, and the low-dimensional evaluation results can be used as new abstract features to replace the original high-dimensional parking data $x_t$, providing a more concise and information-rich feature input for subsequent graph modeling. 

\subsection{ParkingRank Graph Attention Networks}
Graph coarsening, as the current mainstream graph dimensionality reduction technique, aims to overcome the huge computational obstacles faced by large-scale graph data when processing, extracting and analyzing. Typically, the input for graph coarsening is the graph adjacency matrix \cite{20}. When dealing with complex structures such as urban parking graphs, the traditional adjacency matrix constructed from topological relationships is obviously difficult to adequately capture the complexity and diversity of the real parking landscape. In fact, drivers do not only consider the proximity of parking lots in the region when making parking decisions, but also comprehensively compare the distance of regional parking lots, real-time space occupancy rate, openness to the public, and charging situation, and other factors. Therefore, before performing the task of urban parking graph coarsening, we need a high-quality parking graph that can truly reflect the preference of parking behavior and adapt to the dynamic change of parking demand.

GAT \cite{43} are an advanced spatial-based graph neural network methodology, with the core advantage of redefining the aggregation of node features. Unlike GCN, GAT introduces an attention mechanism, allowing the model to dynamically allocate different weights based on the importance of each neighboring node. In contrast, GCN typically assigns equal weights to all neighboring nodes for information aggregation. This strategy enables GAT to more effectively capture spatial correlations within road networks and selectively emphasize nodes that are more critical to the current task. This feature of GAT is particularly important in the application of urban parking network prediction. However, since GAT often uses MLP or cosine similarity to calculate the degree of association between nodes, conventional methods lack interpretability when applied to nodes such as parking lots, which have specific attributes. These methods often fail to provide an intuitive understanding of the interactions between parking lots.

To mitigate the black-box issue encountered by GATs in analyzing parking graphs, we integrate the parking spatiotemporal transfer matrix \cite{2} into the computation of attention coefficients within GATs. This matrix meticulously accounts for the spatiotemporal evolution of parking lot features, encompassing real-time occupancy rates, service capabilities, and spatial connections, thereby facilitating a dynamic representation of parking cruising behavior. This nuanced incorporation allows for a richer, more detailed understanding of parking dynamics, effectively translating the raw data into actionable insights.  The formulation of this transfer matrix $A_{trf}$ is presented in Equation 3.

\begin{equation}
A_{\text {trf }}=\left\{\begin{array}{c}
q_i, \text { when } i=j \\
w_{i, j}\left(1-\mathrm{q}_j\right) P R_i, \text { when } i \neq j
\end{array}\right.,
\end{equation}
where each element $\alpha_{i,j}^{\text{trf}}$  of $A_{\text{trf}} \in \mathbb{R}^{N \times N}$ represents the probability of a vehicle heading to another parking lot $j$ when it finds parking lot $i$ is full. $q_i$ represents the real-time occupancy rate of parking lot $i$, which also indicates the probability of a vehicle choosing to stay in parking lot i based on the parking situation. $w_{i,j}$ is the reciprocal of the normalized distance $A_{i,j}$ between parking lot i and parking lot j. $PR_i$ is the normalized service capacity of parking lot $i$.

By introducing the parking spatiotemporal transfer matrix, the GAT model can dynamically consider multiple factors, including the spatial relationships between parking lots, real-time occupancy rates, service capacities, and other relevant factors, when assigning attention weights. This mechanism enables the model to more accurately capture the spatiotemporal dependencies of nodes in complex parking networks and respond in real-time to fluctuations in parking demand. Through this approach, the model not only identifies the most critical nodes for the current prediction task but also flexibly adjusts its focus on different parking lots, optimizing the model's representation of the parking network's complexity and enhancing its performance in downstream prediction tasks. We believe that the improved graph attention mechanism will help construct a parking network that fully reflects the complex spatiotemporal correlations of parking lots. The overall framework of our proposed PRGAT is shown in Fig. 1:

\begin{figure*}
 \centering
 \includegraphics[width=1\linewidth]{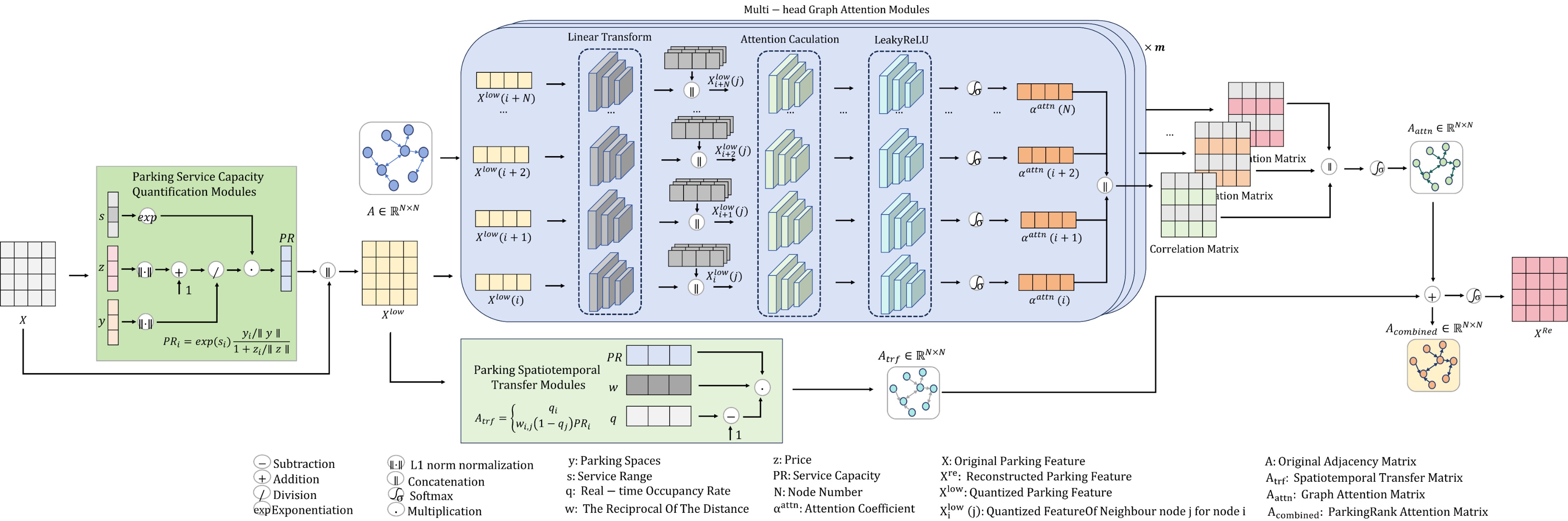}
 \caption{The structure of PRGAT.}
 \label{fig1}
\end{figure*}

In this paper, the input to PRGAT consists of the original adjacency matrix \(A\) based on Euclidean distance and the new parking data \(X^{\text{low}} \in \mathbb{R}^{T \times N \times F^{\text{low}}}\), which is composed of the service capacity \(PR_i\), the real-time occupancy rate \(q_i\), and latitude and longitude coordinates. The output is the updated parking data \(X^{\text{re}} \in \mathbb{R}^{T \times N \times F^{\text{re}}}\). Here, \(F^{\text{low}}\) represents the new feature dimensions after quantification and extraction, while \(F^{\text{re}}\) denotes the new feature dimensions relearned by PRGAT.

The algorithm initially applies a linear transformation to each parking lot node's features \(x_t^{\text{low}}(i)\) using a learnable weight matrix \(W \in \mathbb{R}^{F^{\text{re}} \times F^{\text{low}}}\) to enhance the node's expressive capability. Subsequently, it employs an attention mechanism $\mathcal{T}$ on the set of nodes to compute the attention coefficient \(e_{i,j}\) between node \(i\) and node \(j\). This procedure can be encapsulated by Formula 4, wherein the attention mechanism might be a function that signifies the correlation between two objects, like cosine similarity or a Multilayer Perceptron (MLP), and \(\|\) represents the vector concatenation operation.

\begin{equation}
e_{i, j}=\mathcal{T}\left[W x_t^{\text {low }}(i) \| W x_t^{\text {low }}(j)\right],
\label{eq4}
\end{equation}

To capture local topological information and enhance computational efficiency, a masking mechanism and normalization operation have been introduced. The attention coefficients \(e_{i,j}\) are confined within the first-order neighborhood of the nodes, enabling each node to concentrate solely on its directly connected neighboring nodes and disregard the other nodes in the graph, ultimately yielding the attention matrix \(A_{\text{attn}} \in \mathbb{R}^{N \times N}\). See Formula 5 for details, where LeakyReLU represents the activation function.

\begin{equation}
    A_{\text {attn }}(i, j)=\alpha_{i, j}^{\text {attn }}=\frac{\exp \left(\operatorname{LeakyReLU}\left(e_{i, j}\right)\right)}{\sum_{k \in N_i} \exp \left(\operatorname{LeakyReLU}\left(e_{i, k}\right)\right)},
    \label{eq5}
\end{equation}

Moreover, the masking mechanism ingeniously reflects the factors influencing the parking decision-making process. Drivers, when choosing a parking lot, tend to compare a specific parking lot with its adjacent ones, rather than conducting pairwise comparisons among all parking lots. Such a masking mechanism not only enhances the model's sensitivity to local associations but also aligns more closely with the behavioral patterns in the actual parking decision-making process.

We input both the parking spatiotemporal transfer matrix \(A_{\text{trf}}\) and the aforementioned attention matrix \(A_{\text{attn}}\) into the softmax activation function simultaneously to derive the ParkingRank attention matrix \(A_{\text{combined}} \in \mathbb{R}^{N \times N}\), as indicated in Equation 6.

\begin{equation}
    A_{\text {combined }}(i, j)=\alpha_{i, j}^{\text {combined }}=\operatorname{softmax}\left(\alpha_{i, j}^{\text {trf }}+\alpha_{i, j}^{\text {attn }}\right),
    \label{eq6}
\end{equation}

This ParkingRank attention matrix not only reflects the dynamic characteristics of the parking graph, capturing the flow trajectories and behavioral patterns of vehicles in urban parking scenarios, but also overcomes the shortcomings of insufficient interpretability of the parking lot relevance matrix computed by traditional graph attention methods.

After obtaining the normalized ParkingRank attention coefficients \(\alpha_{i,j}^{\text{combined}}\), GAT conducts a weighted aggregation of the features of each node \(i\) with its neighbors \(N_i\), thereby producing the final output for each node. This procedure is depicted in Formula 7, where \(\sigma\) denotes the sigmoid activation function.

\begin{equation}
    x_t^{r e}(i)=\sigma\left(\sum_{j \in N_i} \alpha_{i, j}^{\text {combined }} W x_t^{\text {low }}(j)\right),
    \label{eq7}
\end{equation}

The PRGAT we designed is fundamentally a pre-trained model, its primary purpose being to serve as a front-end graph construction module within the overall framework for urban parking prediction. Its loss function is described in Formula 8, and the pseudocode for the graph is provided in Algorithm 1.

\begin{equation}
    L_{G A T}\left(X^{\text {low }}, X^{\text {re }}\right)=\frac{1}{T} \sum_{i=t-T+1}^t\left\|X^{\text {low }}(i)-X^{\text {re }}(i)\right\|^2,
    \label{eq8}
\end{equation}

\begin{algorithm}
\caption{ParkingRank Graph Attention Networks (PRGAT)}
\renewcommand{\algorithmicrequire}{\textbf{Input:}}
\renewcommand{\algorithmicensure}{\textbf{Output:}}
\begin{algorithmic}[1]
\REQUIRE $A \in \mathbb{R}^{N \times N}, X \in \mathbb{R}^{T \times N \times F}$
\ENSURE $A_{\text{combined}} \in \mathbb{R}^{N \times N}, {X}^{\text{re}} \in \mathbb{R}^{T \times N \times F_{\text{re}}}$
\STATE $W \leftarrow \text{MLP}(X, F')$
\STATE $a \leftarrow \text{Correlation}(2 \times F, F')$
\STATE $A_{\text{attn}} \leftarrow \text{Zeros}(N, N)$
\STATE $A_{\text{trf}} \leftarrow \text{Zeros}(N, N)$
\FOR{$i \leftarrow 1 \text{ to } N$}
    \STATE $s_i, y_i, z_i \leftarrow X_i[:,\{\text{service range, parking capacity, price}\}]$
    \STATE $PR_i \leftarrow \exp \left(s_i\right) \frac{y_i /\|y\|}{1+z_i /\|z\|}$
    \STATE $X_i^{\text{low}} \leftarrow \text{Concat}\left(X_i[:,\{\text{all}\} \backslash\{\text{service range},\right.$
    \STATE \quad \quad $\text{parking capacity, price} \left.\}], P R_i\right)$
\ENDFOR
\FOR{$i \leftarrow 1 \text{ to } N$}
    \STATE $q_i \leftarrow \frac{X_i[:,\{\text { occupied parking spaces }\}]}{X_i[:,\{\text { parking capacity }\}]}$
    \FOR{$j \leftarrow 1 \text{ to } N$}
        \STATE $w_{i, j} \leftarrow \text{Normalized}(A(i, j))$
        \STATE $\alpha_{i, j}^{\text{trf}} \leftarrow w_{i, j} \ast (1 - q_i) \ast PR_i$
        \STATE $e_{i, j} \leftarrow \text{Concat}(W{X}^{\text{low}}(i), W{X}^{\text{low}}(j))$
        \STATE $\alpha_{i, j}^{\text{ReLU}} \leftarrow \text{LeakyReLU}(a^T \ast e_{i, j})$
        \STATE $\alpha_{i, j}^{\text{Masked}} \leftarrow \text{MaskedFill}(\alpha_{i, j}^{\text{ReLU}}, A(i, j))$
        \STATE $\alpha_{i, j}^{\text{attn}} \leftarrow \text{Normalized}(\alpha_{i, j}^{\text{Masked}})$
        \STATE $A_{\text{combined}}(i, j) \leftarrow \text{softmax}(\alpha_{i, j}^{\text{attn}} + \alpha_{i, j}^{\text{trf}})$
    \ENDFOR
\ENDFOR
\STATE ${X}^{\text{re}} \leftarrow \sigma(A_{\text{combined}} W{X}^{\text{low}})$
\STATE \textbf{return} $A_{\text{combined}}, {X}^{\text{re}}$
\end{algorithmic}
\end{algorithm}

\subsection{Parking Graph Coarsening}
Define the coarsened parking graph \(G_c = (V_c, E_c, A_c)\) is the result of coarsening the original adjacency matrix $A$, which is a smaller weighted graph. \(V_c\) denotes a set of disjoint hypernodes, which covers all nodes in the original graph V, where each hypernode is formed by the aggregation of some of the nodes in the original graph, i.e., \(v_i^c = \{v_1, v_2, \ldots, v_n\} \in \mathbb{R}^N\). \(A_c \in \mathbb{R}^{N_c \times N_c}\) is the adjacency matrix of the coarsened graph, \(N_c = |V_c|\).
 
This document employs the SGC algorithm\cite{21}, which utilizes spectral distance (SD) to demonstrate that the coarsened network retains spatial features similar to those of the original network. This concept is illustrated in the equation below:

\begin{equation}
SD\left(G, G_c\right)=\left\|\lambda-\lambda_c\right\|_1=\sum_{i=1}^N\left|\lambda(i)-\lambda_c(i)\right|,
    \label{eq9}
\end{equation}

where the vectors \( \lambda \) and \( \lambda_c \) represent the eigenvalues of the Laplacian matrices of the original graph \( G \) and the coarsened graph \( G_c \), respectively. The spectral distance \( SD(G, G_c) \) is considered sufficiently small if it is less than \( \delta \) (where \( \delta \) is a very small number). Only when this condition is met do the eigenvalues and eigenvectors of the two graphs exhibit similarity in the spectral domain. The calculated spectral distance can then substantiate that the coarsened graph significantly preserves the attributes of the original graph during the coarsening process. Thus, it can be inferred that the spatial structure of the coarsened network remains similar to that of the original network. The key steps of the SGC algorithm are introduced below.

The algorithm initiates by inputting the ParkingRank attention matrix \(A_{\text{combined}}\) and the coarsening ratio \(n = \frac{N_c}{N}\). It computes the Laplacian matrix \(\mathcal{L}\) and, through matrix decomposition, obtains the first \(N_c\) and the last \(N-N_c+1\) Laplacian eigenvalues \(\lambda\) and Laplacian eigenvectors \(u\), for capturing both the local and global information of the parking graph. This procedure is illustrated in Formulas 10 and 11 , wherein \(\mathcal{L}\) signifies the normalized Laplacian matrix, with \(I_N\) and \(D\) being the identity matrix and the degree matrix, respectively.

\begin{equation}
\mathcal{L}=I_N-D^{-1 / 2} A_{\text {combined }} D^{-1 / 2} ,
\label{eq10}
\end{equation}
\vspace{-20pt}
\begin{equation}
\mathcal{L} u=\lambda D u ,
\label{eq11}
\end{equation}
Then, the algorithm starts to iterate different eigenvalues intervals, performs parking clustering on the internal eigenvectors $u$ respectively, divides the parking lots with similar features into a parking hypernode, and generates a preliminary coarsened graph $G_c$ based on the clustering results, the clustering principle is shown in Equation 12.

\begin{equation}
    d_c(i, j)=\frac{\sum_{i=1, j=1}^n u_i u_j}{\sqrt{\sum_{i=1}^n\left(u_i\right)^2} \sqrt{\sum_{j=1}^n\left(u_j\right)^2}},
    \label{eq12}
\end{equation}

The Laplacian eigenvectors $u$ computed from the ParkingRank attention matrix $A_{combined}$ has rich parking graph topology information as well as parking lot node feature information, which can provide more accurate and intuitive parking lot node characterization for the coarsening process.

The process continuously calculates the distance \(SD\) between the Laplacian eigenvectors of the coarsened graph \(G_c\) and those of the original graph, opting for the coarsening outcome with the minimum error. Finally, the coarsened graph \(G_c = (V_c, E_c, A_c)\) along with the corresponding index matrix \(index\) are returned. This procedure is depicted in Formula 13, with the coarsening specifics provided in Algorithm 2.

\begin{equation}
    SD\left(A_{\text {combined }}, A_c\right)=\left\|d_{\text {combined }}(i)-d_c(i)\right\|_1,
    \label{eq13}
\end{equation}

\begin{algorithm}
\caption{Parking Graph Coarsening}
\renewcommand{\algorithmicrequire}{\textbf{Input:}}
\renewcommand{\algorithmicensure}{\textbf{Output:}}
\begin{algorithmic}[1]
\REQUIRE $A_{\text{combined}} \in \mathbb{R}^{N \times N}, n$
\ENSURE $A_c \in \mathbb{R}^{N_c \times N_c}, \text{index}$
\STATE $\text { index } \leftarrow[] \text {, temp } \leftarrow 0$
\STATE $N_c \gets \text{CoarsenedNodesNum}(n, N)$
\STATE $\mathcal{L} \gets \text{NormalizedLaplacian}(A_{\text{combined}})$
\STATE $\lambda, u \gets \text{EigenValueAndVector}(L, N_c)$
\FOR{$i \gets 1 \text{ to } N$}
    \FOR{$j \gets 1 \text{ to Neighbor }(i)$}
        \STATE $d_c(i, j)=\frac{\sum_{i=1, j=1}^n u_i u_j}{\sqrt{\sum_{i=1}^n\left(u_i\right)^2} \sqrt{\sum_{j=1}^n\left(u_j\right)^2}}$
        \STATE $\text { index }[\text { temp }] \leftarrow \text { index }[ \text { ++temp }] \cup\{\mathrm{i}, \mathrm{j}\}$
        \FOR{$q \gets 1 \text{ to Neighbor } (temp)$}
            \STATE $A_c^{\text {temp} }($temp,$q)+=\sum_{x \in i, j} A_{\text {combined }}(x, q)$
        \ENDFOR
    \ENDFOR
\ENDFOR
\FOR{$i \gets 1 \text{ to } N$}
    \FOR{$j \gets 1 \text{ to } N$}
        \STATE $A_c \gets \arg\min \text{SD}(A_{\text{combined}}(i, j), A_c(i, j))$
    \ENDFOR
\ENDFOR
\STATE \textbf{return} $A_c, \text{index}$
\end{algorithmic}
\end{algorithm}

\subsection{Prediction Framework Based On Coarsened Parking Graphs}
Since the coarsening process of the parking graph described above does not downscale the parking lot node features, the feature data corresponding to the parking hypernodes \(X^{\text{combined}} \in \mathbb{R}^{T \times N \times F^{\text{low}}}\)remains a splicing operation between the merged parking lot data, which is formalized as follows:

\begin{equation}
    x_t^{\text {combined }}(i)=\sum_{j \in v_i^c} x_t^{\text {low }}(i) \| x_t^{\text {low }}(j),
    \label{eq14}
\end{equation}

If it is simply used as the feature input for the parking prediction models, it will not significantly improve the overall framework training efficiency, and even lose the intrinsic connection between the nodes. In this regard, we consider adopting the symmetric TCN-AE, which is because: 
\begin{itemize}
    \item TCN is able to mine the intrinsic laws behind the parking time-series data itself \cite{42}, such as the tidal characteristics, which helps in the compression and reconstruction of the parking data.
    \item The encoder is able to embed the high-dimensional sparse parking data into the low-dimensional dense tensor form, which reduces the computational overhead of the training model.
    \item The decoder is able to achieve an approximate lossless reduction and can reconstruct the spatial structure of the original parking graph.  
\end{itemize}

\begin{figure}[H]
 \centering
 % \includesvg[width=1\linewidth]{pics/fig2.svg}
 \includegraphics[width=1\linewidth]{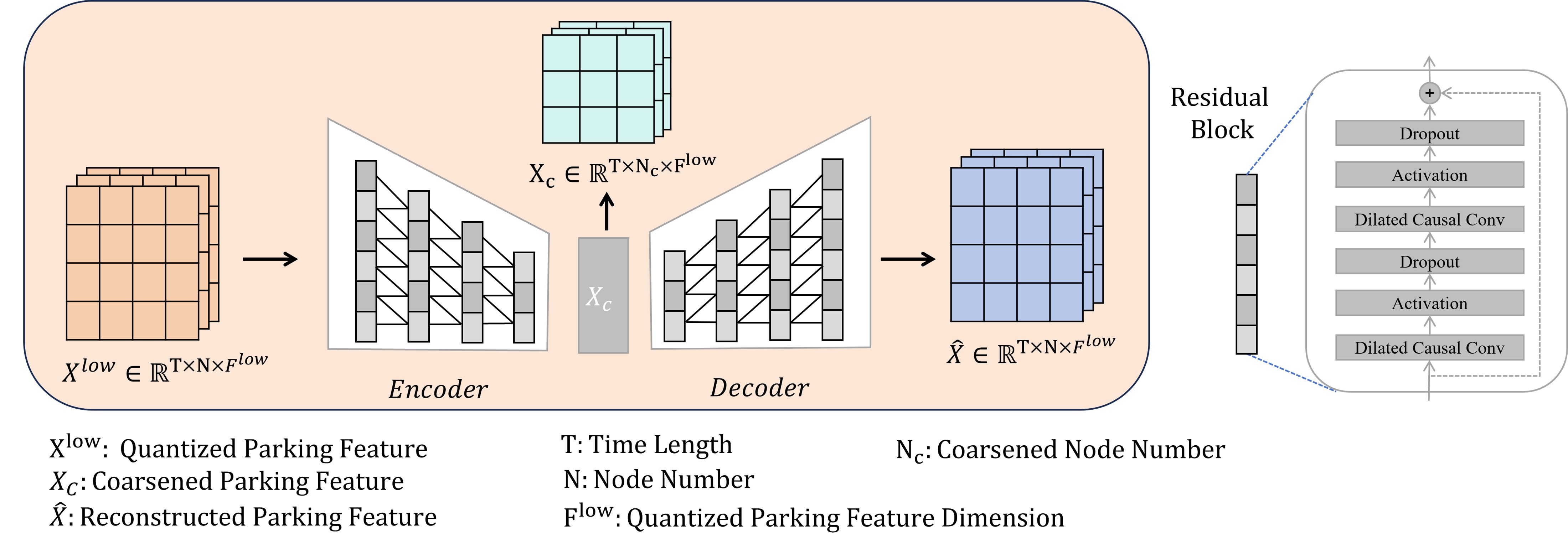}
 \caption{The structure of TCN-AE.}
 \label{fig2}
\end{figure}

To maintain the consistency of the prediction results between the coarsened and the original parking graphs, we train a set of symmetric TCN-based AEs for each parking hypernode, utilizing the index matrix generated by the graph coarsening module. TCNs typically consist of multiple layers of dilated causal convolution, with each layer performing one-dimensional convolution. Moreover, to enhance the network's expressive capacity, residual connections are often added following multiple convolutional layers, organizing these layers into several residual blocks. Each residual block primarily includes two dilated causal convolution layers, interspersed with normalization, activation, and dropout operations, before the output is added to the input of the next residual block. This design not only optimizes the flow of information but also enhances the model's ability to capture features of parking data effectively.

The procedure is detailed as follows: As depicted in Fig. 2, firstly, the parking data \(X^{\text{low}} \in \mathbb{R}^{T \times N \times F^{\text{low}}}\) undergoes encoding with the residual blocks within the encoder\cite{40}, as indicated in Formula 15. Owing to the reduction in the number of features as the quantity of residual blocks increases, it allows the feature data \(X^c \in \mathbb{R}^{T \times N_c \times F^{\text{low}}}\) corresponding to the coarsened graph to be represented within a low-dimensional feature space.

\begin{equation}
    X^{\text{c}}(i)=\left(X^{\text {low }} * f\right)(i)=\sum_{j=0}^{k-1} f(j) X^{\text {low }}(i-j),
    \label{eq15}
\end{equation}
where $\sum_{j=0}^{k-1} f(j)$ denotes the filter $f$ of length $k$ and $*$ denotes the convolution operation. This formula ensures that only information prior to the current time step is considered and future information is not computed.

\begin{figure}[H]
 \centering
 % \includesvg[width=1\linewidth]{pics/fig2.svg}
 \includegraphics[width=1\linewidth]{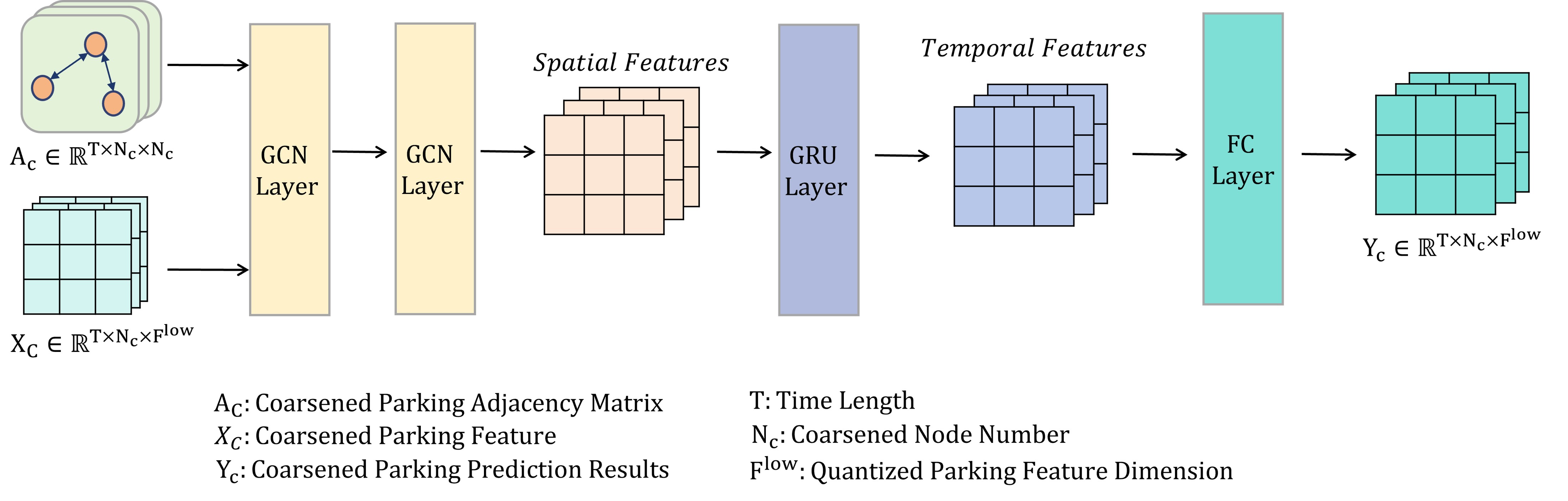}
 \caption{The structure of T-GCN.}
 \label{fig3}
\end{figure}

Next, we input the  adjacency matrix $A_c$ of coarsened graph and feature matrix $X^c$ into a spatiotemporal graph convolutional model. This paper takes T-GCN\cite{30} as an example, as shown in Fig. 3,  which captures the spatiotemporal dependencies of the parking coarsened graph through 2 layers of GCN and 1 layer of GRU, finally obtaining the coarsened graph's predicted results \(Y_c \in \mathbb{R}^{T \times N_c \times F^{\text{low}}}\) through a linear transformation, as seen in Formulas 16, 17, 18, 19, and 20, where $W$ and $b$ represent the weights and biases of T-GCN, respectively, and \(u_t\), \(r_t\), \(c_t\), \(h_t\) denote the reset gate, update gate, candidate hidden state, and current hidden state at time $t$,respectively. $gc\left(\cdot\right)$ denotes the graph convolution operation, and tanh is the activation function.

\begin{equation}
u_t=\sigma\left(W_u \cdot\left[g c\left(X^c, A_c\right), h_{t-1}\right]+b_u\right),
\label{eq16}
\end{equation}
\vspace{-15pt}
\begin{equation}
r_t=\sigma\left(W_r \cdot\left[gc\left(X^c, A_c\right), h_{t-1}\right]+b_r\right) ,
\label{eq17}
\end{equation}
\vspace{-15pt}
\begin{equation}
c_t=\tanh \left(W_c \cdot\left[g c\left(X^c, A_c\right),\left(r_t, h_{t-1}\right)\right]+b_c\right),
\label{eq18}
\end{equation}
\vspace{-15pt}
\begin{equation}
h_t=u_t * h_{t-1}+\left(1-u_t\right) * c_t,
\label{eq19}
\end{equation}
\vspace{-15pt}
\begin{equation}
Y_c=\operatorname{softmax}\left(W_y \cdot h_t+b_y\right),
\label{eq20}
\end{equation}

Due to the fact that the dimensions of the coarsened graph's predicted results \(Y_c\) do not align with the size of the original parking graph, these results are not inherently interpretable. To address this, decoding is necessary. Based on the prediction length, we opt for a particular pre-trained decoder, inputting the prediction of each hypernode to reconstruct the original parking graph's prediction results \(\hat{Y} \in \mathbb{R}^{T \times N \times F^{\text{low}}}\), thereby fulfilling the prediction task for the entire urban parking graph. The decoding procedure is depicted in Formula 21. \(f^{-1}\) denotes the filters within the decoder, distinct from the encoding phase, as the feature count in the decoder's residual blocks grows with the increase in residual blocks, aiming to reconstruct parking data within a high-dimensional feature space.

\begin{equation}
    \hat{Y}(i)=\left(Y_c * f^{-1}\right)(i)=\sum_{s=0}^{T-1} \sum_{i=0}^{k-1} f^{-1}(i) Y_c(i),
    \label{eq21}
\end{equation}

The loss functions for the pre-trained TCN-AE and the parking prediction model utilize Mean Squared Error (MSE) and Huber Loss, respectively, as detailed in Formulas 22 and 23. In this context, \(\hat{X} \in \mathbb{R}^{T \times N \times F^{\text{low}}}\) represents the outcome post-reconstruction by the autoencoder, \(Y\) signifies the real observed data, and \(\theta\) represents the threshold parameter that controls MSE.

\begin{equation}
    L_{A E}\left(X^{\text {low }}, \hat{X}\right)=\frac{1}{T^{\prime}} \sum_{i=t-T^{\prime}+1}^t\left\|X^{\text {low }}(i)-\hat{X}(i)\right\|^2 ,
    \label{eq22}
\end{equation}
% \vspace{-10pt}
\begin{equation}
    L_{\text {predict }}(Y, \hat{Y})=\left\{\begin{array}{c}
\frac{1}{2}(Y-\hat{Y})^2,|Y-\hat{Y}| \leq \theta \\
\theta|Y-\hat{Y}|-\frac{1}{2} \theta^2, \text { otherwise }
\end{array}\right.,
\label{eq23}
\end{equation}

The overall parking prediction framework of this paper is shown in Fig. 4.
\begin{figure*}[]
 \centering
 \includegraphics[width=0.75\linewidth]{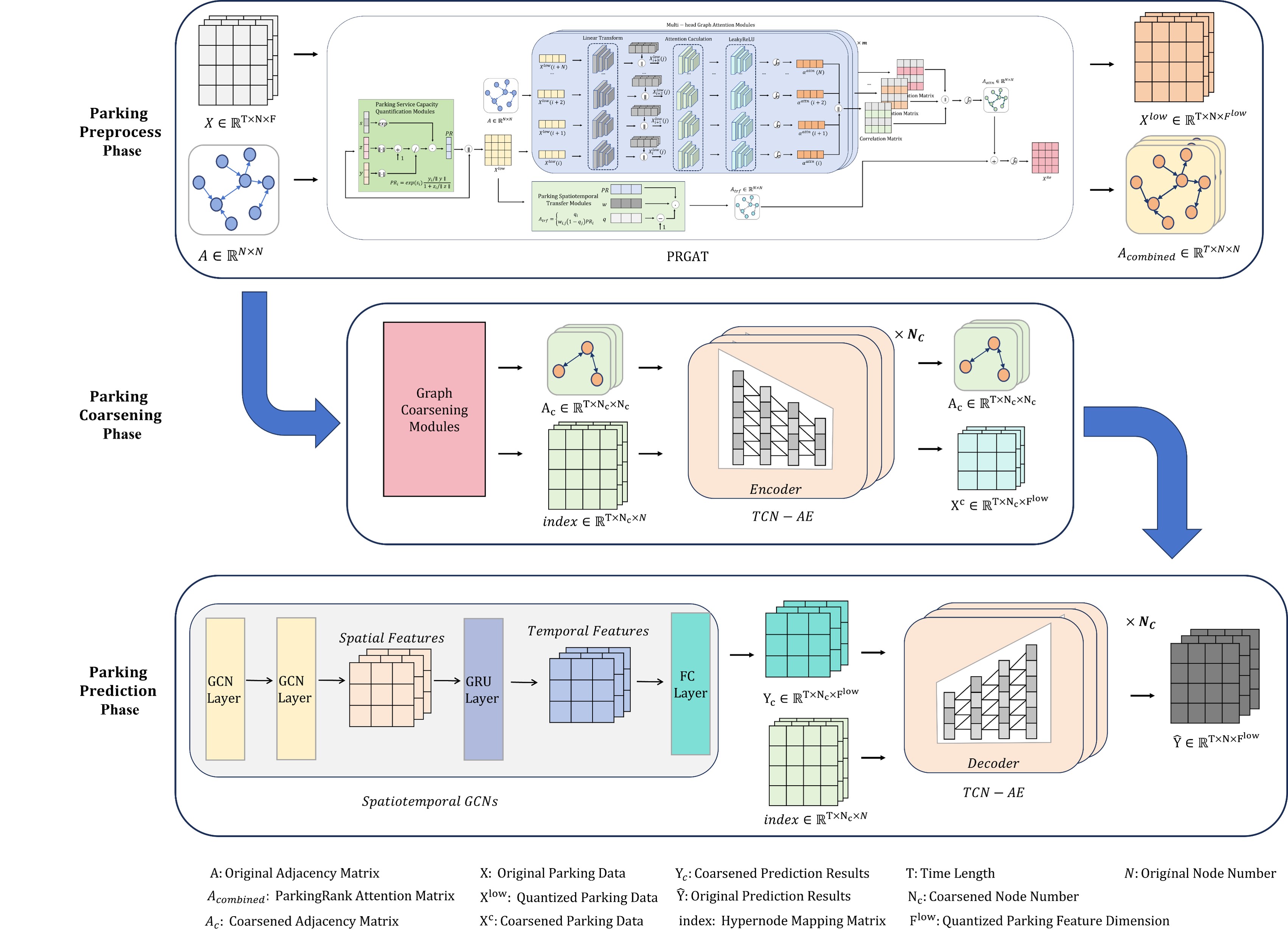}
 \caption{The structure of Urban Parking Prediction Framework.}
 \label{fig4}
\end{figure*}

\section{EXPERIMENTS}
\subsection{Experiment Setup}
This paper selected 9000 parking lots within five districts including Bao'an and Luohu in Shenzhen as research subjects. The parking data recorded for six months starting from June 1, 2016, was sampled at 15-minute intervals. These datasets encompassed key information such as the geographic coordinates, open status, pricing details, and real-time occupancy of each parking lot. To ensure data quality, we conducted deduplication and visualization-based outlier detection operations on the raw parking data. Additionally, to maintain consistency in data processing, all datasets underwent normalization before analysis. Finally, we partitioned the dataset chronologically into training set (70\%), validation set (20\%), and test set (10\%) to ensure the validity and reliability of the experiments. Specifically, the distribution of parking lots in Bao'an District was visually presented through Fig. 5, and specific details of the parking lot data are referred to in Table \Rmnum{1}.

\begin{figure}
 \centering
 \includegraphics[width=1\linewidth]{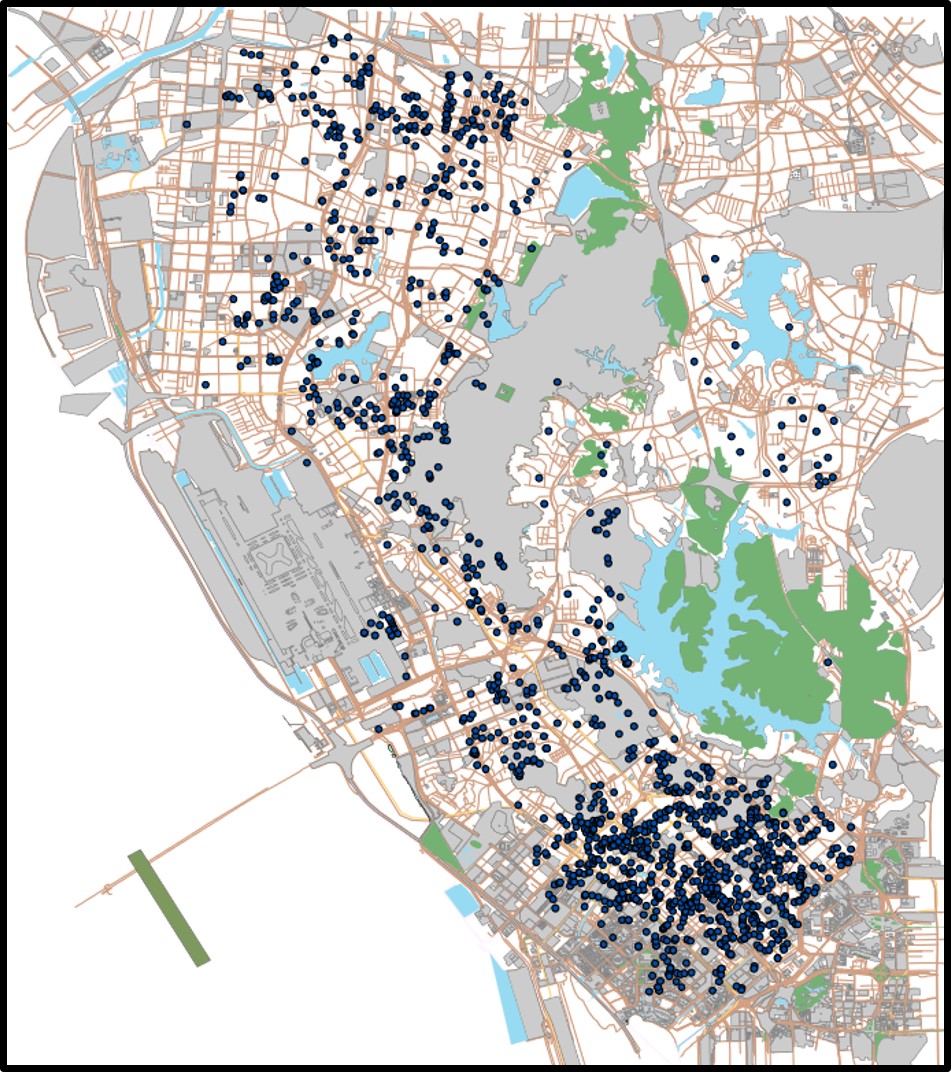}
 \caption{Distribution map of parking lots in Bao'an District.}
 \label{fig5}
\end{figure}

% \usepackage{graphicx}
\begin{table}[h]
\label{table1}
\renewcommand{\arraystretch}{1.5}
\centering
\caption{Description of Shenzhen Parking Lot Dataset}
\begin{tabular}{lllll}
\hline
District & \#Nodes & Start Time & Granularity & Time Steps \\ \hline
Bao'an    & 1660    & 2016/6/1   & 15min       & 17,280     \\
Luohu    & 1730    & 2016/6/1   & 15min       & 17,280     \\
Futian   & 1806    & 2016/6/1   & 15min       & 17,280     \\
Longgang & 1473    & 2016/6/1   & 15min       & 17,280     \\
Longhua  & 2531    & 2016/6/1   & 15min       & 17,280     \\ \hline
\end{tabular}%
\end{table}

To evaluate the performance of our method, we selected several commonly used evaluation metrics in regression tasks: MAE , RMSE, and MAPE. The expressions for these metrics are as follows:
\begin{equation}
    \label{eq24}
    \begin{aligned}
        \text{MAE} &= \frac{1}{TN} \sum_{t=1}^{T} \sum_{i=1}^{N} |y_t^i - \hat{y}_t^i| \\
        \text{RMSE} &= \sqrt{\frac{1}{TN} \sum_{t=1}^{T} \sum_{i=1}^{N} (y_t^i - \hat{y}_t^i)^2} \\
        \text{MAPE} &= \frac{100\%}{TN} \sum_{t=1}^{T} \sum_{i=1}^{N} \left| \frac{y_t^i - \hat{y}_t^i}{y_t^i} \right|
    \end{aligned}
\end{equation}
where \(\hat{y}_t^i\) and \(y_t^i\) respectively represent the predicted result and the ground truth for the \(i\)-th parking lot at time \(t\). \(T\) denotes the length of the time sample, and \(N\) is the total number of parking lot nodes.

In the experimental parameter design, we use data from the past 12 time steps to predict data for the next 1, 2, and 4 steps, which corresponds to using the past 3 hours of parking traffic to predict the parking traffic for the next 15 minutes, 30 minutes, and 60 minutes, respectively. Additionally, to evaluate the performance of our proposed urban parking prediction framework, we selected several widely-used baseline models in the field of parking prediction, as well as some recent popular graph reduction methods for comparison:
\begin{itemize}
\item T-GCN\cite{32}: A classical model for traffic flow prediction that combines Graph Convolutional Networks (GCNs) and Gated Recurrent Units (GRU) to capture the spatiotemporal correlations in traffic data, effectively modeling both spatial dependencies between traffic nodes and temporal patterns within the traffic time series.
\item STGCN\cite{44}: Utilizes multiple ST-Conv blocks that integrate temporal and spatial convolutions to model multi-scale traffic networks, proven to effectively capture comprehensive spatiotemporal correlations.
\item STSGCN\cite{45}: By stacking multiple STSGCL blocks for synchronous spatiotemporal modeling, it can effectively capture complex local spatiotemporal traffic correlations.
\item SparRL\cite{35}: A universal and effective graph sparsification framework implemented through deep reinforcement learning, capable of flexibly adapting to various sparsification objectives.
\end{itemize}

For the training of different models, we used a V100 GPU for development and incorporated early stopping and dynamic learning rate adjustment strategies to optimize training performance. To ensure completeness and transparency, all model configurations and training-related hyperparameter settings are detailed in Table \Rmnum{2}. In these configurations, the structural parameters of each model were kept at their default settings, while other hyperparameters, such as learning rate and weight decay, were selected using grid search.

\begin{table*}[]
\normalsize
\label{table2}
\centering
\caption{Description of Model Configurations and Training Parameters}
\centering
\setlength{\tabcolsep}{3pt}
\renewcommand{\arraystretch}{1.2}
\resizebox{\textwidth}{!}{%
\begin{tabular}{ccccccccc}
\hline
Method & \multicolumn{2}{c}{Configuration} & Batch Size & Learning Rate & Optimizer & Loss Function & Weight Decay & Patience (/Epoch) \\ \hline
\multirow{2}{*}{PRGAT} & Number of Attention Heads & Feature Dimension per Head & \multirow{2}{*}{64} & \multirow{2}{*}{1e-4} & \multirow{2}{*}{Adam} & \multirow{2}{*}{MSE} & \multirow{2}{*}{1e-4} & \multirow{2}{*}{100} \\ \cline{2-3}
 & 8 & 128 &  &  &  &  &  &  \\ \hline
\multirow{2}{*}{SGC} & \multicolumn{2}{c}{Threshhold \( \lambda \)} & \multirow{2}{*}{-} & \multirow{2}{*}{-} & \multirow{2}{*}{-} & \multirow{2}{*}{-} & \multirow{2}{*}{-} & \multirow{2}{*}{-} \\ \cline{2-3}
 & \multicolumn{2}{c}{1e-8} &  &  &  &  &  &  \\ \hline
\multirow{2}{*}{SparRL} & \multicolumn{2}{c}{Maximum number of neighbors to pay attention to} & \multirow{2}{*}{64} & \multirow{2}{*}{1e-4} & \multirow{2}{*}{Adam} & \multirow{2}{*}{Huber Loss} & \multirow{2}{*}{1e-4} & \multirow{2}{*}{500} \\ \cline{2-3}
 & \multicolumn{2}{c}{64} &  &  &  &  &  &  \\ \hline
\multirow{2}{*}{TGCN} & \multicolumn{2}{c}{GRU Hidden Units} & \multirow{2}{*}{64} & \multirow{2}{*}{1e-5} & \multirow{2}{*}{Adam} & \multirow{2}{*}{Huber Loss} & \multirow{2}{*}{1e-4} & \multirow{2}{*}{200} \\ \cline{2-3}
 & \multicolumn{2}{c}{100} &  &  &  &  &  &  \\ \hline
\multirow{2}{*}{STGCN} & Graph Convolution Dimension & Temporal Convolution Dimension & \multirow{2}{*}{64} & \multirow{2}{*}{1e-5} & \multirow{2}{*}{Adam} & \multirow{2}{*}{Huber Loss} & \multirow{2}{*}{1e-4} & \multirow{2}{*}{200} \\ \cline{2-3}
 & 16 & 64 &  &  &  &  &  &  \\ \hline
\multirow{2}{*}{STSGCN} & GCNs per Module & Spatiotemporal GCNs Layers(STSGCL) & \multirow{2}{*}{64} & \multirow{2}{*}{1e-5} & \multirow{2}{*}{Adam} & \multirow{2}{*}{Huber Loss} & \multirow{2}{*}{1e-4} & \multirow{2}{*}{200} \\ \cline{2-3}
 & 3 & 4 &  &  &  &  &  &  \\ \hline
\multirow{2}{*}{TCN-AE} & TCN Dilation Rates & Filter Count and Kernel Size & \multirow{2}{*}{64} & \multirow{2}{*}{1e-4} & \multirow{2}{*}{Adam} & \multirow{2}{*}{MSE} & \multirow{2}{*}{1e-4} & \multirow{2}{*}{100} \\ \cline{2-3}
 & (1,2,4,8,16) & 20 &  &  &  &  &  &  \\ \hline
\end{tabular}%
}
\end{table*}

\subsection{Modeling Real Scenes}
To demonstrate the graph construction process of the PRGAT method, we conducted an experiment using real parking lot data from Bao'an District in Shenzhen. In Bao'an District, the number of parking lots exceeds 1,000, leading to the number of reachable edges between parking lots reaching the order of millions. Therefore, to simplify the analysis while ensuring the practicality of the results, we selected a more central CBD area within the district as the study area, which contains over 600 parking lots. Additionally, we set a spatial distance threshold of 500 meters; distances between any parking lots exceeding this threshold were considered unreachable to align with real parking scenarios.

\begin{figure}[H]
 \centering
 % \includesvg[width=1\linewidth]{pics/fig4.svg}   
 \includegraphics[width=1\linewidth]{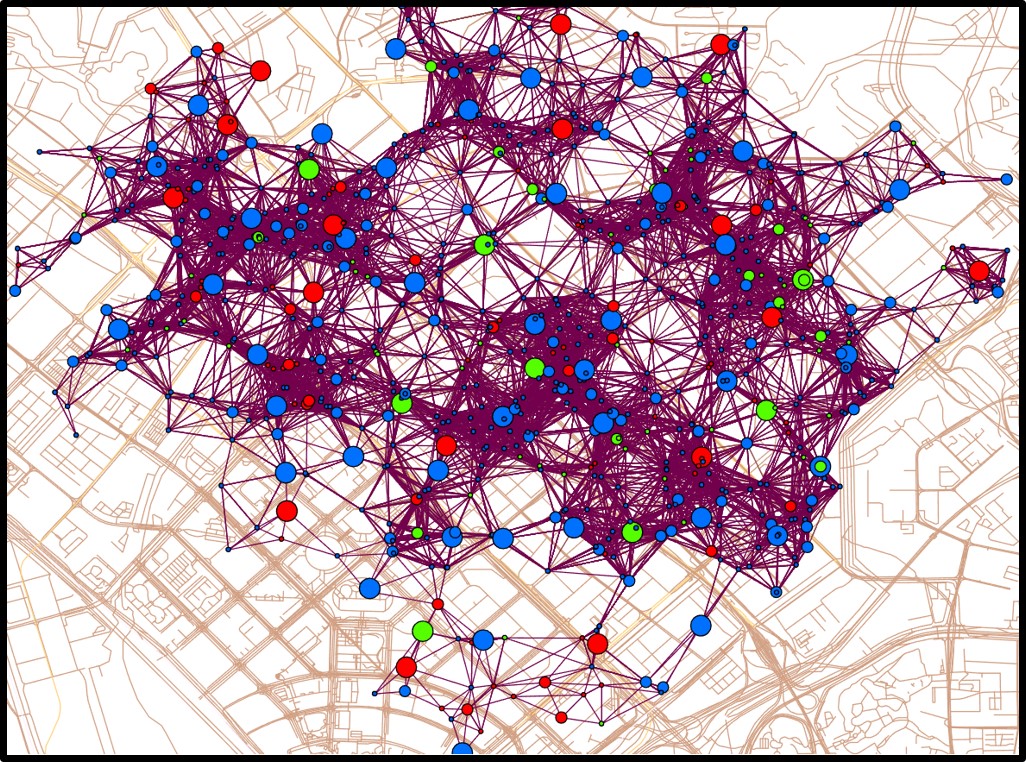}
 \caption{The parking graph of the core area of Bao'an, encompassing over 600 parking facilities.}
 \label{fig6}
 % \vspace{-2em}
\end{figure}

Shown in Fig. 6 is a visualization of the parking graph constructed using the PRGAT model for the Baoan core area around evening during the week, where the different sizes of circles represent the service capacity of the parking lots, with larger circles implying stronger service capacity and more popularity among drivers. Unlike the traditional spatial adjacency matrix, the dynamic adjacency matrix generated by PRGAT can intuitively reflect the real-time importance of each parking lot, which helps the downstream prediction model to accurately determine the key parking lots in each time period, thus effectively identifying the places with strong or weak parking demand. Meanwhile, the graph composed of these parking lots with strong service capacity essentially depicts the parking backbone network of the city in different time periods, and this dynamic parking backbone network reflects the core areas and hotspots of the city's parking demand, which provides an important perspective for understanding the city's parking behavior pattern.

\begin{figure*}
 \centering
 % \includesvg[width=1\linewidth]{pics/fig5.svg}   
 \includegraphics[width=1\linewidth]{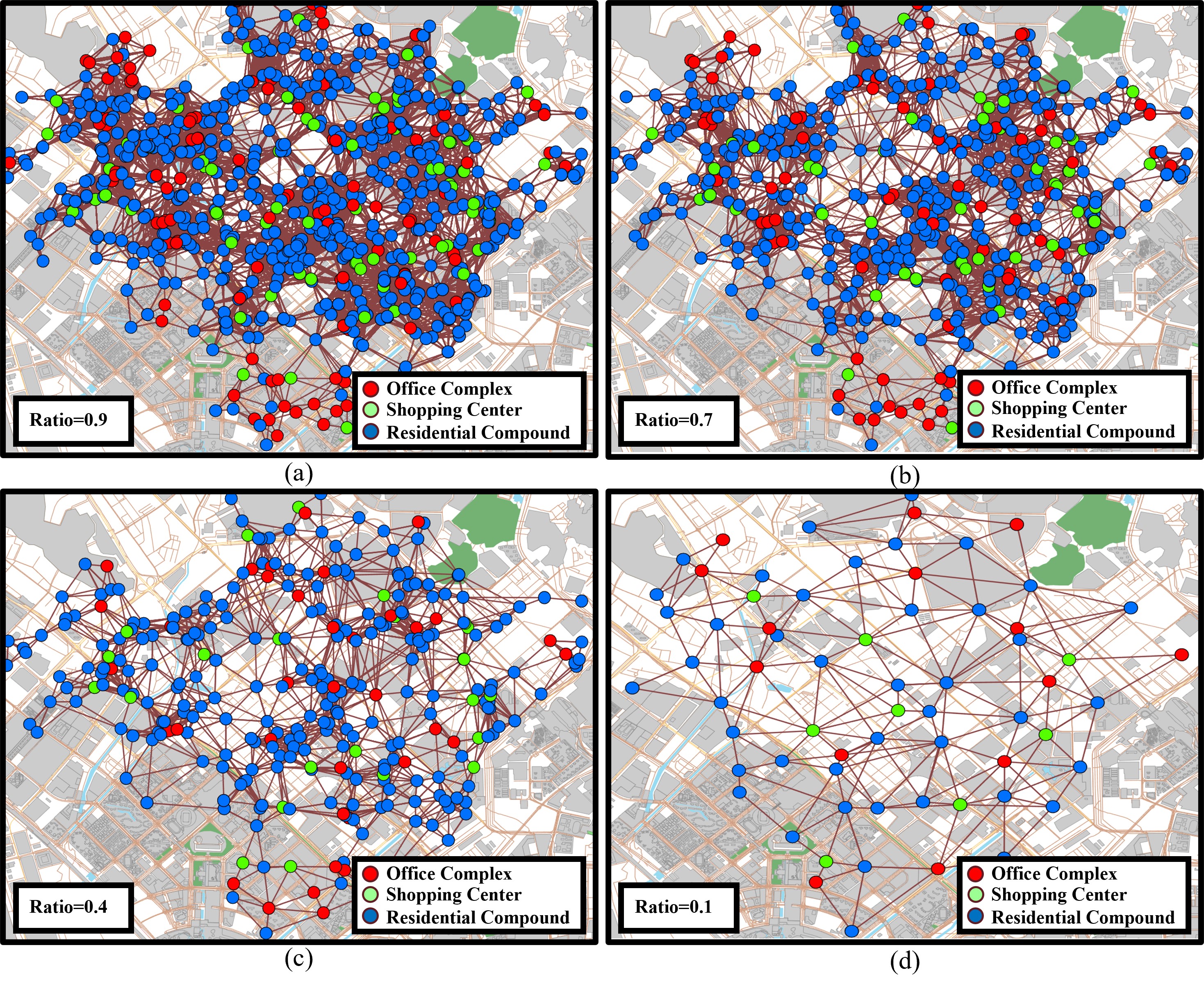}
 \caption{Distribution of parking lots in the core area of Bao'an under different coarsening granularities: (a) Coarsening ratio = 0.9 (b) Coarsening ratio = 0.7 (c) Coarsening ratio = 0.4 (d) Coarsening ratio = 0.1.}
 \label{fig5}
\end{figure*}

Fig. 7 depicts the distribution of parking lots in the core area of Bao'an under different coarsening ratio (coarsened vertices/original graph vertices). By adjusting the coarsening ratio, we can clearly see significant changes in the structure and density of the parking graph: parking lots with stronger service capacities gradually absorb those with weaker capacities, thus forming larger parking hypernodes with enhanced comprehensive service capabilities. As the ratio of coarsening is progressively reduced, the backbone of the urban parking graph also becomes clearer. However, this coarsening process also poses certain challenges: a simplified parking graph means the loss of some important parking information, which could affect the downstream model's ability to capture details, especially in areas highly sensitive to parking demand or where demand changes rapidly.

\subsection{Selection of Coarsening Ratio}
Considering the dual role of the coarsening ratio in the speed of model convergence and model prediction accuracy, we set up control experiments with coarsening ratio of 0.2, 0.3, 0.4, 0.5, 0.6, 0.7, 0.8, 0.9, and without coarsening (coarsening ratio of 1), to study the impact of different coarsening ratio on the prediction of parking occupation rate during the evening rush hour. The prediction results are shown in the Figure 8, where dashed and solid lines respectively represent the impact of different coarsening ratio on model prediction accuracy and convergence speed.

As shown in Fig. 8, when the coarsening ratio decreases from 1 to 0.6, the error growth rate of the model predictions remains relatively stable. This indicates that within this range, key features are effectively preserved, and the model primarily loses redundant and secondary parking information, whose absence has limited impact on overall prediction accuracy. However, as the coarsening ratio is further reduced, prediction error rises sharply. For instance, at a coarsening ratio of 0.2, calculation errors of 77.2\%, 64.2\%, and 61.8\% are introduced across different model configurations. This suggests that crucial parking information, including the highly variable characteristics of certain parking lots, has been stripped away during the coarsening process, which significantly degrades model performance.

\begin{figure}[H]
 \centering
 % \includesvg[width=1\linewidth]{pics/fig6.svg} 
 \includegraphics[width=1\linewidth]{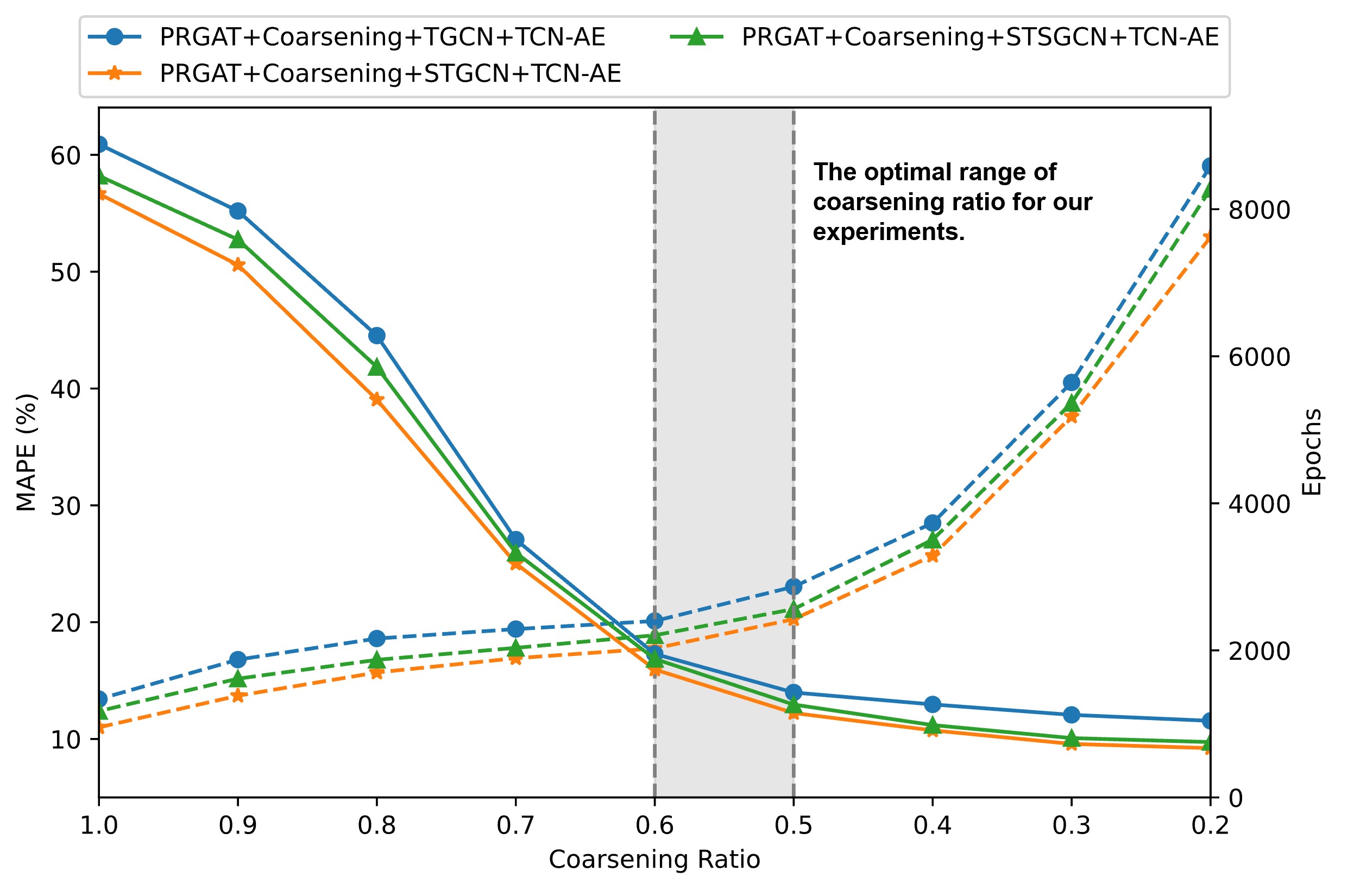}
 \caption{Impact of Different Coarsening Ratio on Prediction Performance.}
 \label{fig8}
\end{figure}

It's noteworthy that the curve of model training duration presents a different trend. Initially, as the coarsening ratio reduces, training time decreases rapidly, demonstrating that coarsening can effectively alleviate the model’s computational burden. However, once the coarsening ratio reaches 0.6, the reduction in training time becomes more gradual, suggesting that further coarsening does not yield significant efficiency gains. This shift indicates that the coarsening process has started merging nodes of parking lots with distinct characteristics, resulting in the loss of critical detail information that was previously observable. Under these conditions, the model’s ability to differentiate between variations in parking lot service capacities gradually deteriorates, making it less efficient in learning the complex features of urban parking networks in a lower-information setting. This, in turn, impacts its prediction accuracy and generalization performance.

The two sets of experiments indicate that the ideal coarsening ratio should be within the range of 0.5 to 0.6. Coarsening within this interval manages to strike a balance between simplifying model complexity and retaining key feature information, which is particularly crucial when dealing with large-scale urban parking graphs, as these often contain a vast number of parking lot nodes and edges. Moreover, this phenomenon mirrors the real-world situation where approximately 50\%-60\% of parking lots make the primary contribution to urban parking demand. These parking lots are typically the ones with the highest usage frequency, optimal locations, and best service capabilities. However, it should be noted that the coarsening rate range we provide is merely idealized. The specific choice of coarsening rate must be dynamically adjusted based on the structure of the urban parking network and characteristics of parking data (such as parking frequency, peak periods) using adaptive deep learning models.

 % \vspace{-9pt}
 % \vspace{-0.55em}
\subsection{Quantitative Results}
To explore the predictive performance of the urban parking prediction framework we proposed, we conducted predictions on the future 15 minutes, 30 minutes, and 60 minutes parking occupancy rates of various parking lots in Bao'an District during weekday evening rush hours using three baseline models, with the coarsening ratio set to 0.6. The corresponding prediction results are presented in Table \Rmnum {3}. 

Our experimental results demonstrate that, whether for short-term or long-term predictions, models using PRGAT as a pre-construction unit for graphs almost always outperform traditional parking prediction methods across all metrics. Specifically, taking TGCN as an example, compared to the default adjacency matrix, the ParkingRank attention matrix created by PRGAT achieves an average reduction of about 46.8\% in MAE. Furthermore, compared to the traditional GAT mechanism, it also reduces RMSE by about 20.5\%. This significant improvement in performance is attributed to the comprehensive considerations PRGAT takes in constructing dynamic parking graphs.  Unlike traditional methods that focus solely on the spatial relationships between parking lots or on a single feature like available parking spaces, PRGAT also incorporates other critical information including real-time service capabilities such as the available capacity, level of accessibility, and cost of parking lots. This approach results in a graph that is not just a simple representation of spatial proximity but a richer, more accurate depiction of actual parking behavior patterns, greatly enhancing the model's understanding and predictive ability in complex parking scenarios.

% \usepackage{multirow}
% \usepackage{graphicx}b
\begin{table*}[]
 \vspace{-1.0em}
\normalsize
\label{table3}
\centering
\caption{Comparison of Predicted Parking Lot Occupancy Rates in Bao'an District for the Next Hour}
\centering
\resizebox{\textwidth}{!}{%
\begin{tabular}{lllllllllll}
\hline
\multicolumn{1}{c}{\multirow{2}{*}{Method}} & \multicolumn{3}{c}{MAE}                             & \multicolumn{3}{c}{RMSE}                              & \multicolumn{3}{c}{MAPE(\%)}                     & \multicolumn{1}{c}{\multirow{2}{*}{Epoch}} \\ \cline{2-10}
\multicolumn{1}{c}{}                        & 15min           & 30min           & 60min           & 15min           & 30min            & 60min            & 15min          & 30min          & 60min          & \multicolumn{1}{c}{}                       \\ \hline
Default+TGCN                                & 7.933           & 8.7224          & 11.3431         & 11.2128         & 12.3413          & 15.1087          & 13.82          & 16.23          & 20.59          & 834                                        \\
GAT+TGCN                                    & 7.7224          & 8.3431          & 10.2926         & 10.0689         & 11.4238          & 12.3675          & 12.74          & 14.41          & 19.82          & 839                                        \\
PRGAT+TGCN                                  & 4.2631          & 4.719           & 5.8398          & 7.4103          & 8.7885           & 10.7526          & 9.89           & 10.38          & 13.09          & 844                                        \\
Default+Coarsening+TGCN+AE                  & 8.0153          & 8.9607          & 11.8285         & 11.4731         & 13.455           & 16.0598          & 14.31          & 17.25          & 20.77          & 655                                        \\
Default+Sparsification+TGCN+AE              & 9.9726          & 11.1673         & 13.6201         & 13.4071         & 14.757           & 18.6382          & 14.22          & 17.54          & 22.96          & 827                                        \\
PRGAT+Sparsification+TGCN+TCN-AE            & 7.851           & 9.0267          & 11.5006         & 12.4467         & 12.6782          & 15.483           & 13.36          & 15.28          & 18.64          & 846                                        \\
PRGAT+Coarsening+TGCN+TCN-AE                & \textbf{4.6552} & \textbf{5.209}  & \textbf{7.3209} & \textbf{7.8395} & \textbf{9.4337}  & \textbf{10.9448} & \textbf{10.24} & \textbf{10.53} & \textbf{13.49} & \textbf{558}                               \\ \hline
Default+STGCN                               & 4.728           & 5.5534          & 9.598           & 10.7798         & 11.7946          & 14.9729          & 10.68          & 16.47          & 19.12          & 673                                        \\
GAT+STGCN                                   & 3.7947          & 4.3721          & 7.6845          & 7.44            & 10.9601          & 12.4731          & 9.42           & 14.2           & 16.93          & 645                                        \\
PRGAT+STGCN                                 & 2.159           & 2.7601          & 4.328           & 4.893           & 6.4853           & 10.4475          & 7.63           & 10.65          & 11.74          & 639                                        \\
Default+Coarsening+STGCN+AE                 & 5.4295          & 7.6965          & 10.2398         & 11.4307         & 12.9623          & 16.0431          & 11.74          & 17.83          & 20.01          & 568                                        \\
Default+Sparsification+STGCN+AE             & 6.5246          & 7.9378          & 11.0445         & 12.6193         & 14.7207          & 17.8087          & 13.47          & 18.55          & 20.79          & 651                                        \\
PRGAT+Sparsification+STGCN+TCN-AE           & 5.0214          & 6.5779          & 9.538           & 10.6814         & 12.5294          & 15.7111          & 11.43          & 14.84          & 17.17          & 662                                        \\
PRGAT+Coarsening+STGCN+TCN-AE               & \textbf{2.1977} & \textbf{3.3915} & \textbf{5.3604} & \textbf{4.9548} & \textbf{7.8741}  & \textbf{10.9849} & \textbf{8.65}  & \textbf{11.6}  & \textbf{12.31} & \textbf{476}                               \\ \hline
Default+STSGCN                              & 6.5194          & 6.5358          & 9.5476          & 10.8984         & 14.4152          & 17.1354          & 12.19          & 14.37          & 16.33          & 788                                        \\
GAT+STSGCN                                  & 5.6569          & 6.1912          & 8.1355          & 9.8521          & 13.534           & 16.7885          & 10.71          & 11.75          & 12.49          & 752                                        \\
PRGAT+STSGCN                                & 4.1645          & 4.6743          & 7.6569          & 7.1038          & 9.4388           & 11.8712          & 7.85           & 8.22           & 8.89           & 769                                        \\
Default+Coarsening+STSGCN+AE                & 7.3511          & 7.8178          & 11.0491         & 11.4534         & 14.6645          & 18.5779          & 12.67          & 15.03          & 16.82          & 661                                        \\
Default+Sparsification+STSGCN+AE            & 8.1708          & 10.319          & 12.4703         & 13.3065         & 15.2423          & 19.1752          & 14.11          & 17.33          & 19.8           & 776                                        \\
PRGAT+Sparsification+STSGCN+TCN-AE          & 6.6163          & 7.656           & 9.7815          & 11.8712         & 13.9123          & 18.7807          & 12.82          & 13.94          & 17.25          & 740                                        \\
PRGAT+Coarsening+STSGCN+TCN-AE              & \textbf{4.6079} & \textbf{4.9813} & \textbf{8.093}  & \textbf{7.9218} & \textbf{10.0797} & \textbf{12.9521} & \textbf{8.72}  & \textbf{9.31}  & \textbf{10.09} & \textbf{556}                               \\ \hline
\end{tabular}%
}
\end{table*}

In terms of enhancing model training efficiency, compared to baseline methods, our framework achieves an average improvement of approximately 30.5\%, further confirming the efficiency of our framework in handling complex parking prediction tasks. Furthermore, in the choice of graph dimensionality reduction techniques, graph coarsening outperforms graph sparsification in reducing model training duration. Graph sparsification primarily reduces the scale of the graph by decreasing the number of edges, but since the dimensionality of node features remains unchanged, its impact on training efficiency is not significant. Additionally, the loss of crucial connectivity information between parking lots can severely affect the predictive accuracy of downstream models. In contrast, graph coarsening, by merging nodes, not only reduces the scale of the parking network but also simplifies the model's computational complexity while preserving key structural information, thereby aiding in maintaining the accuracy of the model's predictions.

Notably, we optimized the AE by incorporating TCN to improve the processing of parking time series data, achieving near-lossless compression and restoration of large-scale parking data. Compared to previous work \cite{46}, this approach effectively keeps the MAPE within 12.6\%, further increasing the accuracy of predictions.

It is important to recognize that although the application of coarsening techniques can improve model training efficiency, it inevitably introduces some level of error. Hence, our aim is to maximize training efficiency while striving to preserve prediction accuracy as much as possible. As the graph scale expands, our framework not only demonstrates a distinct advantage in prediction accuracy but also reveals increasingly significant differences in efficiency. This is especially evident when handling large-scale, dynamically changing complex urban parking graphs, showcasing the exceptional performance and practical value of our framework.
 \vspace{-1.0em}
\subsection{Scalability Results}
To verify whether our proposed framework can adapt to parking graphs of different scales, we simulate the expansion of the parking graph scale by gradually increasing the number of parking lots. In the experiments, the coarsening ratio is set to 0.6, with the prediction of parking occupancy rates for the next 15 minutes during the evening rush hour as the evaluation task. This approach allows us to observe changes in the model's prediction accuracy and convergence speed on parking graphs of various sizes.
 % \vspace{-1.0em}
Synthesizing Fig. 9(a) and 9(b), it can be observed that when the scale of the parking graph is small, our framework shows relatively conservative improvements in terms of accuracy and efficiency, with an average accuracy improvement of about 11.1\% and a training efficiency improvement of about 23.7\%. Importantly, as the scale of the graph expands, the performance advantages of our prediction framework become particularly pronounced. Unlike the baseline models, which exhibit a higher error growth rate and exponential increase in iteration costs, our framework demonstrates a stable growth trend, significantly surpassing traditional methods in terms of error control and training efficiency.

\begin{figure}
 \centering
 % \includesvg[width=1\linewidth]{pics/fig7.svg} 
 \includegraphics[width=1\linewidth]{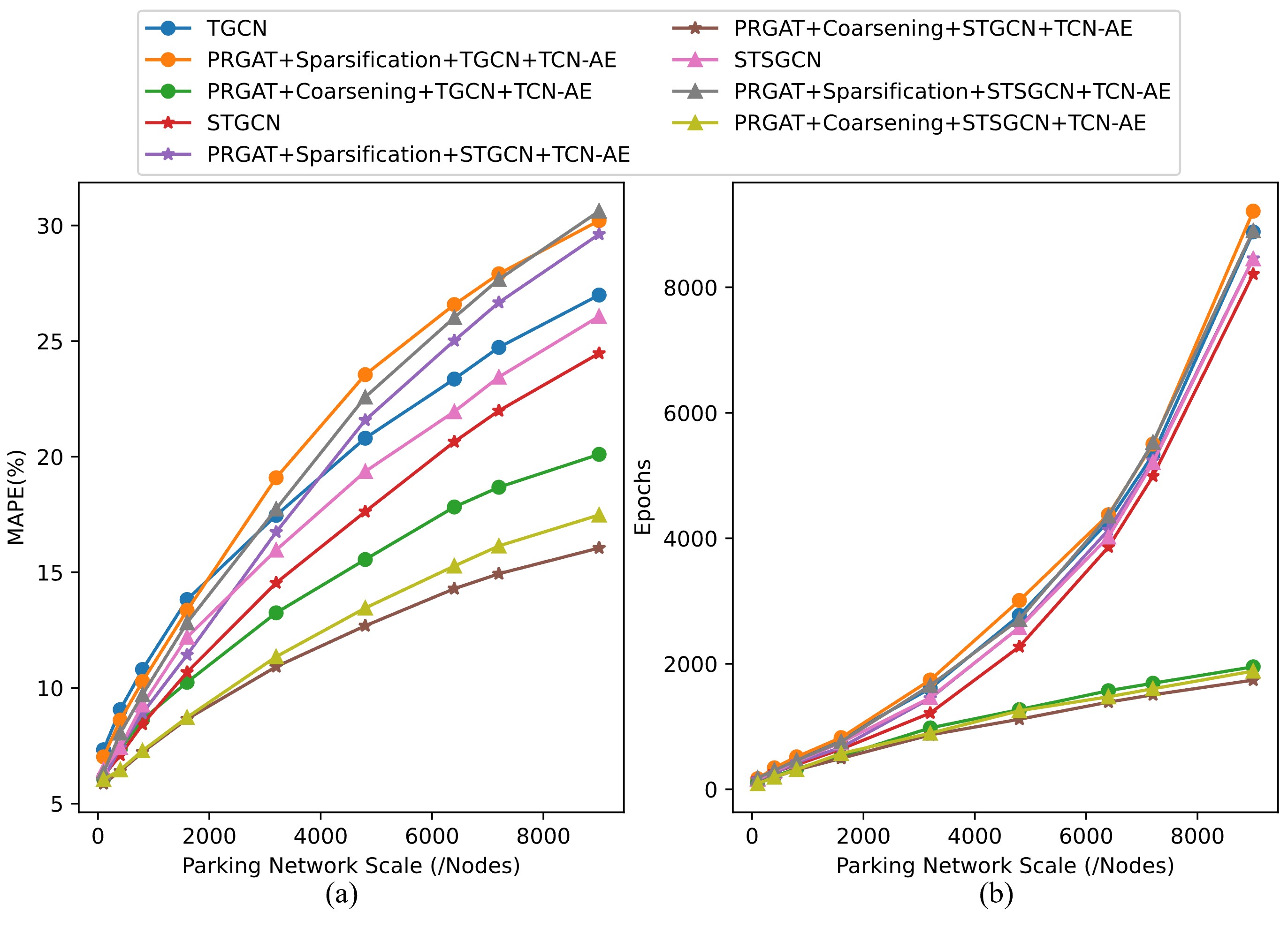}
 \caption{Comparison of Prediction Performance for Future 15-Minute Parking Lot Occupancy Rates Across Parking Graphs of Various Scales (a) MAPE (b) Convergence Situation.}
 \label{fig9}
\end{figure}

This performance advantage is mainly attributed to the efficient algorithms and model structures adopted within our framework, such as the PRGAT and TCN-AE . The former effectively captures key features of parking data, while the use of the latter significantly enhances the model's capability to extract and analyze parking lot features. This ensures that the prediction framework can retain critical information from the original parking graph to a great extent, even after coarsening processing. In contrast, traditional parking prediction techniques, even when combined with graph sparsification, still face issues with suboptimal prediction performance and training speeds. This is due to graph sparsification often losing significant parking detail information as the network scale increases, severely impacting predictive performance. Additionally, since this technique does not involve dimension reduction of node features, the growth in training efficiency is similar to baseline models that do not employ any graph dimension reduction techniques.

In addressing the challenges of large-scale urban parking prediction, our framework optimizes data flow and information processing, effectively managing complex and voluminous parking data. This not only ensures high accuracy in model predictions but also significantly reduces training times, thereby meeting the needs for timeliness and accuracy. Such capabilities enable our framework to quickly and accurately respond to dynamic changes in urban parking, providing real-time predictive results.

\subsection{Robustness Analysis}
To validate the robustness of our framework under different parking patterns and regional layouts, we further refined the experimental dataset and conducted robustness analysis on the subdivided datasets. Specifically, using the parking data of Bao'an District as the base, we divided the dataset into three smaller subsets based on differences in parking lot service coverage. These subsets correspond to regions primarily dominated by shopping centers, office complexes, and areas with a relatively balanced distribution of parking lot types. Each subset contains approximately 200 parking lots, as detailed in Table \Rmnum {4}. In the experimental setup, we set the prediction step to 1, the coarsening rate to 0.6, and used MAE and MAPE as the main evaluation metrics.

\begin{table}[H]
\label{table4}
\renewcommand{\arraystretch}{1.2}
\centering
\caption{Dataset Settings for Different Parking Types}
\begin{tabular}{cccc}
\hline
Dataset Type & \begin{tabular}[c]{@{}c@{}}Shopping\\  Center\\ Parking Lots \\ (Count)\end{tabular} & \begin{tabular}[c]{@{}c@{}}Office \\ Complex\\ Parking Lots\\ (Count)\end{tabular} & \begin{tabular}[c]{@{}c@{}}Residential\\  Compound\\ Parking Lots\\ (Count)\end{tabular} \\ \hline
Commercial Dataset & 133 & 30 & 37 \\
Office Dataset & 34 & 122 & 44 \\
Balanced Dataset & 72 & 65 & 63 \\ \hline
\end{tabular}
\end{table}

As shown in Table \Rmnum {5}, the prediction results for different types of parking datasets demonstrate that our framework achieves higher prediction accuracy on datasets with distinct regional characteristics compared to parking networks with evenly distributed parking lot types. Specifically, in parking networks dominated by commercial area parking lots, our framework improves prediction accuracy by 11.59\% in terms of MAE. Similarly, for parking networks dominated by office area parking lots, the MAE prediction error is reduced by an average of 13.18\%. In terms of MAPE, our framework achieves an average improvement of 15.06\% in prediction performance on these two types of parking networks with highly concentrated service characteristics.

% \usepackage{multirow}
\begin{table}[H]
\label{table5}
\footnotesize
\setlength{\tabcolsep}{3pt}
\renewcommand{\arraystretch}{1.2}
\centering
\caption{Comparison of 15-Minute Ahead Prediction Performance Across Different Types of Parking Datasets}
\begin{tabular}{ccccccc}
\hline
\multirow{2}{*}{Datasets} & \multicolumn{2}{c}{\begin{tabular}[c]{@{}c@{}}PRGAT+\\ Coarsening+\\ TGCN+TCN-AE\end{tabular}} & \multicolumn{2}{c}{\begin{tabular}[c]{@{}c@{}}PRGAT+\\ Coarsening+\\ STGCN+TCN-AE\end{tabular}} & \multicolumn{2}{c}{\begin{tabular}[c]{@{}c@{}}PRGAT+\\ Coarsening+\\ STSGCN+TCN-AE\end{tabular}} \\ \cline{2-7} 
 & MAE & MAPE(\%) & MAE & MAPE(\%) & MAE & MAPE(\%) \\ \hline
\begin{tabular}[c]{@{}c@{}}Commercial\\ Dataset\end{tabular} & 3.7208 & 6.14 & 2.2168 & 5.58 & 4.0399 & 5.63 \\
\begin{tabular}[c]{@{}c@{}}Office\\ Dataset\end{tabular} & 3.5564 & 6.37 & 2.1077 & 5.42 & 4.1988 & 5.80 \\
\begin{tabular}[c]{@{}c@{}}Balanced\\ Dataset\end{tabular} & 4.3507 & 7.44 & 2.4326 & 6.71 & 4.5624 & 6.44 \\ \hline
\end{tabular}
\end{table}

The performance improvement is primarily attributed to the fact that parking networks with distinct regional characteristics often exhibit clearer parking demand and behavior patterns. For example, the demand fluctuations for commercial area parking lots are mainly concentrated on weekends, while the peak demand for office area parking lots typically occurs during morning and evening rush hours. Such parking networks with well-defined overall trends provide a solid foundation for the optimization of the PRGAT module in our framework. Specifically, the characteristics of these networks enable the PRGAT module to accurately identify and enhance the feature representation of critical nodes while weakening the influence of secondary and redundant nodes on model performance. This allows the construction of an adjacency matrix that better reflects actual parking behavior patterns. The adjacency matrix, in turn, provides the downstream spatiotemporal graph convolutional model with clearer learning objectives, thereby improving the model's prediction performance. 

In contrast, for parking networks with evenly distributed types and less pronounced demand characteristics, the correlations between nodes are relatively uniform, which limits the effectiveness of the PRGAT module in constructing the adjacency matrix and results in a smaller improvement in the framework's predictive performance.Through these experimental validations, our framework is better suited for parking networks with distinct regional characteristics, whereas the overall performance improvement is relatively limited for networks with more evenly distributed parking types.

% \vspace{-0.945em}
\subsection{Ablation Results}
To further validate the rationality of each component within the framework, we conducted an ablation study based on the task of predicting the occupancy rate of parking lots in the Bao'an District for the next 15 minutes during the evening rush hour. The ablation experiments were divided into two parts, separately assessing the impact of PRGAT and TCN-AE on three sets of baseline models, with the coarsening ratio set to 0.6 in the experiments.

Fig. 10(a) shows the experimental results of parking predictions using various types of adjacency matrices. Clearly, among all baseline methods, the ParkingRank attention matrix constructed by PRGAT exhibits the best performance, achieving an average reduction in relative error of about 38.3\%. PRGAT, by integrating the attention coefficient matrix with the parking spatiotemporal transition matrix, delves deep into the feature and spatial dependencies between parking lots, effectively restoring real parking demand scenarios. In contrast, traditional static adjacency matrices, built solely on spatial location information, overlook the intrinsic characteristics of parking lots and their interactions with the surrounding environment. Often, they fail to fully capture the complex interactions and dynamic changes between parking lots, limiting the authenticity and dynamism of the scenarios captured by the model, and thus performing poorly in downstream prediction tasks. Moreover, compared to the traditional GAT mechanism, another advantage of PRGAT is that it not only offers a more detailed and dynamic method to represent parking graphs, but it also intuitively reflects which parking lots have stronger comprehensive service capabilities at specific moments, thus becoming a key basis for making parking decisions.

However, a limitation of the PRGAT is that it aggregates node feature information considering only the 2nd-order neighborhood, neglecting the potential connections between distant nodes. This may lead to inadequate capturing of the interactions between remote parking lots during holidays or major events. Therefore, future research exploring and implementing graph convolutional networks that include higher-order neighborhood information will be crucial. This will enable the model to capture a broader range of node interactions, thereby enhancing the accuracy and practicality of the predictions.

\begin{figure}[h]
 \centering
 \includegraphics[width=1\linewidth]{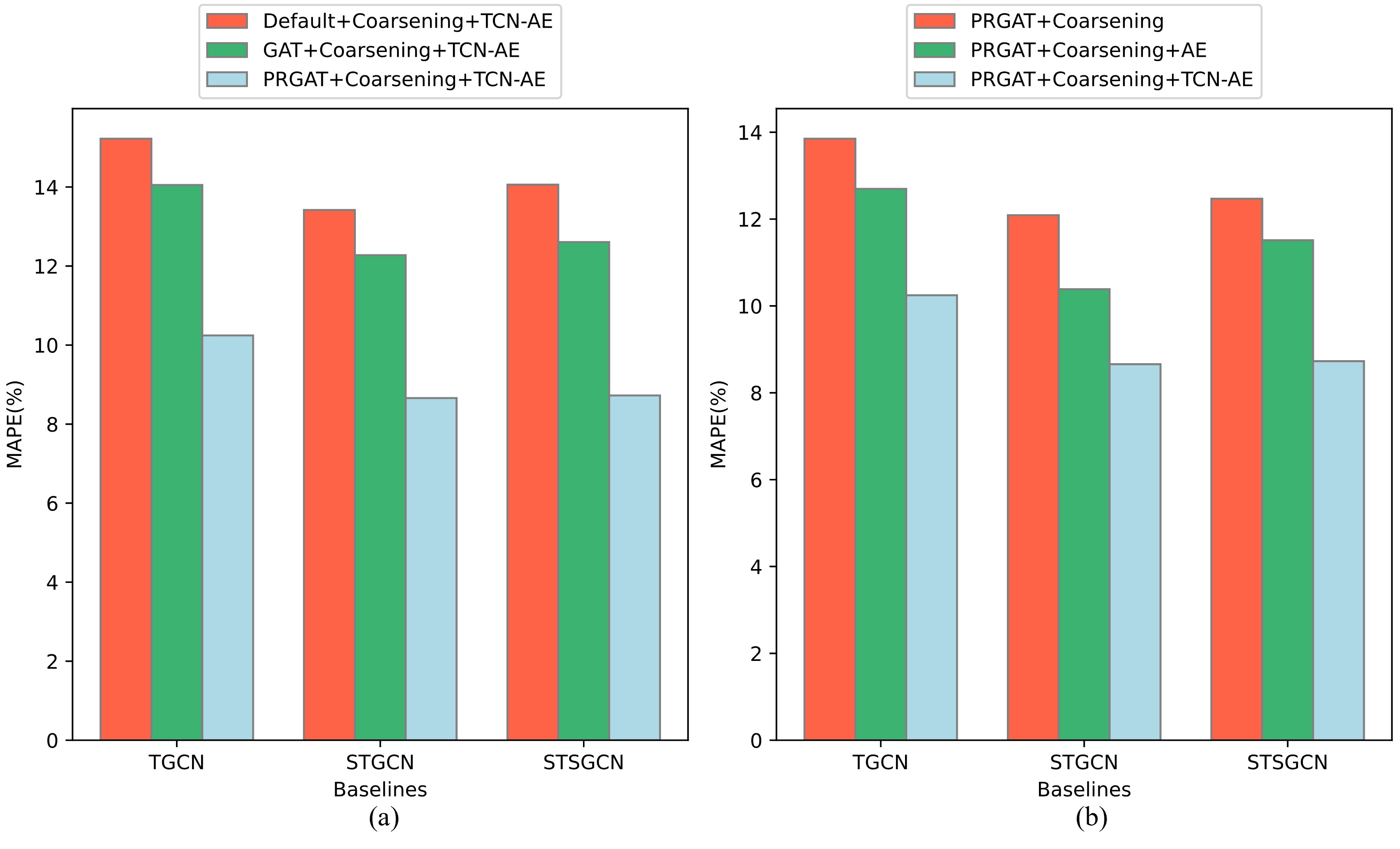}
 \caption{Ablation Study on Future 15-Minute Parking Prediction. (a) Graph Construction (b) Autoencoder.}
 \label{fig10}
\end{figure}

Fig. 10(b) illustrates the effect of the AE on the prediction framework. From the figure, it can be found that not adding the AE leads to an average decrease in prediction accuracy of about 30.1\%. This is because the coarsened parking graph changes its original spatial relationship, and simply splicing a set of vertex features corresponding to each hypernode will lead to redundancy of hypernode information and dimensionality inflation, thus limiting the model's generalization ability. In addition, the loss of the intrinsic connection between nodes will also lead to the model's inability to fully exploit the spatial dependency of the coarsened graph. On the contrary, the approach using autoencoder can selectively extract and reconstruct the key features inside the hypernodes, thus effectively adapting to the spatial changes inside the hypernodes. Meanwhile, we additionally introduce TCN in the encoding-decoding process, which enhances the model's ability to learn the laws behind the dynamic changes of the parking lot view from the time-series data, and is more helpful for the compression and reconstruction of the parking data, which achieves an average 20\% accuracy improvement.

\subsection{Node Error Analysis}
To further validate the contribution of the ParkingRank Graph Attention Module to the accuracy of parking prediction tasks, we conducted a node error comparison experiment. Specifically, we used the original adjacency matrix and the attention matrix generated by the pre-trained ParkingRank Graph Attention Module as inputs to the spatiotemporal graph convolution model and recorded the average prediction error of each parking lot node during training to evaluate the impact of different adjacency matrices on model performance. The experiment utilized data from 100 parking lots in Bao'an District, with T-GCN as the parking prediction model and a prediction step set to 1.

\begin{figure}[h]
 \centering
 \includegraphics[width=1\linewidth]{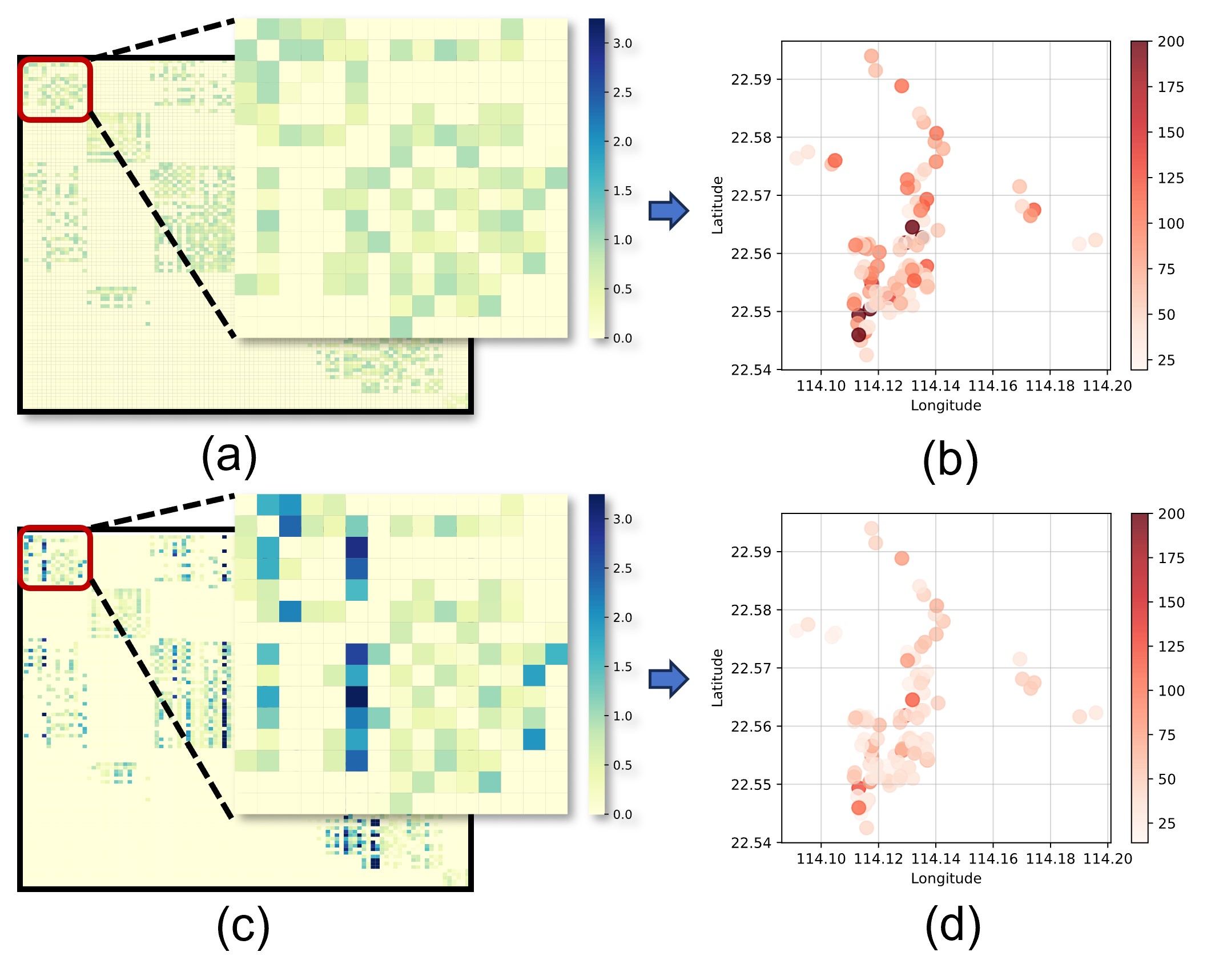}
 \caption{Heatmap Visualization. (a) Original Adjacency Matrix (b) Node Average Prediction Error Distribution under the Original Adjacency Matrix (c) ParkingRank Attention Matrix (d) Node Average Prediction Error Distribution under the ParkingRank Attention Matrix.}
 \label{fig11}
\end{figure}

Fig. 11(a) shows the heatmap visualization of the original adjacency matrix, where each cell represents the spatial relationships between different parking lots. From the figure, it can be observed that the weight distribution is relatively uniform, lacking evident traffic patterns and presenting a certain level of noise. Fig. 11(b) illustrates the prediction error distribution of different parking lot nodes during the training process when the original adjacency matrix is used as input for the spatiotemporal graph convolution model. It is evident that the prediction errors for all parking lot nodes are generally high, indicating that the model struggles to effectively capture the spatiotemporal relationships between parking lots. For certain important parking lots in the area (e.g., those near densely packed commercial zones or office buildings, highlighted in deep red in the figure), the prediction errors are particularly pronounced. This phenomenon further demonstrates that the original adjacency matrix fails to effectively reflect the underlying traffic patterns and spatial features of the parking network, thereby limiting the model's prediction performance.

Fig. 11(c) presents the heatmap visualization of the ParkingRank Attention Adjacency Matrix constructed by the ParkingRank Graph Attention Module. Compared to Fig. 11(a), the weight distribution in Fig. 11(c) is more heterogeneous and structured. Specifically, the distribution characteristics of important parking lots are more evident in Fig. 11(c), as reflected by larger weight accumulations in the corresponding rows or columns, with overall darker colors. This feature indicates that certain parking lots are likely located in commercial centers or other areas with high parking demand, showing stronger associations with nearby parking lots. As a result, distinct cluster-like structures are formed in the corresponding rows or columns of the adjacency matrix. These cluster structures reflect the local correlations between parking lots, effectively revealing the actual characteristics of the parking network and its spatiotemporal interaction patterns.

Fig. 11(d) illustrates the prediction error distribution of different parking lot nodes during the training process when the ParkingRank Attention Matrix is used as input. Compared to Fig. 11(b), where the original adjacency matrix is used, the error levels of all parking lot nodes are significantly reduced. In particular, for nodes at the center of cluster structures (i.e., parking lot nodes with darker colors in Fig. 11(b)), the model effectively reduces the average prediction error by 45.1\% during each training epoch. This result demonstrates that the ParkingRank Attention Matrix enables the spatiotemporal graph convolution model to achieve higher learning efficiency and lower error levels in regions with distinct features. Specifically, the ParkingRank Attention Matrix optimizes the weight distribution of the original adjacency matrix, more accurately reflecting the relationships between important parking lots and their surrounding lots. This refined weight allocation not only strengthens the relational representation of key nodes but also reduces the influence of minor and redundant nodes, allowing the model to learn more realistic parking behavior patterns in high-demand areas, thereby improving the prediction accuracy for these parking lot nodes. In contrast, the original adjacency matrix, with its uniform and unstructured weight distribution, results in concentrated errors in important parking lot areas during training, further hindering the model's overall prediction performance.

\subsection{Comparison of Computational Complexity}
\begin{table*}[]
 \vspace{-1.0em}
 \renewcommand{\arraystretch}{1.5}
\normalsize
\label{table6}
\centering
\caption{Comparison of Time and Space Complexity Across Different Parking Network Scales}
\resizebox{\textwidth}{!}{
\begin{tabular}{ccccccccc}
\hline
\multirow{2}{*}{} & \multicolumn{2}{c}{1800Nodes} & \multicolumn{2}{c}{3600Nodes} & \multicolumn{2}{c}{5400Nodes} & \multicolumn{2}{c}{7200Nodes} \\ \cline{2-9} 
 & FLOPs & MU/Byte & FLOPs & MU/Byte & FLOPs & MU/Byte & FLOPs & MU/Byte \\ \hline
STSGCN & \multicolumn{1}{l}{83.27G} & \multicolumn{1}{l}{297.59M} & \multicolumn{1}{l}{198.54G} & \multicolumn{1}{l}{789.69M} & \multicolumn{1}{l}{277.8G} & \multicolumn{1}{l}{1517.25M} & \multicolumn{1}{l}{395.07G} & \multicolumn{1}{l}{2467.30M} \\ \hline
PRGAT+Coarsening+STSGCN+TCN-AE & \textbf{52.31G} & 350.16M & \textbf{127.29G} & 955.78M & \textbf{191.56G} & 1917.7M & \textbf{274.69G} & 3082.93M \\ \hline
\end{tabular}
}
\end{table*}

To gain a deeper understanding of the time and space complexity of our framework when training on large-scale urban parking networks, we conducted a quantification experiment on computational cost (FLOPs) and memory usage (MU). In the evaluation process, we selected STSGCN as the baseline and set the graph coarsening rate to 0.6. The specific computational complexity comparison results are shown in Table \Rmnum {6}.

As shown in Table \Rmnum {6}, in terms of time complexity, our framework reduces the computational cost of each forward and backward pass by an average of 33.2\% when training on large-scale parking networks. This optimization is primarily attributed to the graph coarsening technique, which compresses the original parking network into a smaller graph while preserving the main structure and spatiotemporal dependencies of the original graph. As a result, the model only needs to compute fewer nodes and edges during each forward and backward pass. In contrast, traditional parking prediction models typically rely on full-graph computation, where each node aggregates information from all its neighboring nodes in each layer. As the graph size increases, this approach leads to a significant increase in computational cost, which severely impacts training efficiency.It is worth noting that, while we introduce the PRGAT, Graph Coarsening, and TCN-AE modules into our framework, which inevitably add extra computation, the impact of these additional computations on the overall training time is negligible and can be considered almost insignificant when compared to traditional models that perform spatiotemporal graph convolution directly on the original parking network scale.

Specifically, the computational cost of the PRGAT module is proportional to the graph size, but through the selective weighting mechanism, the model does not perform equal computations for all neighboring nodes. Instead, it focuses on the neighbors most relevant to the current node, thereby reducing redundant calculations and avoiding unnecessary overhead. The Graph Coarsening module’s computational cost mainly comes from node merging and adjacency matrix updates, but compared to the downstream spatiotemporal GCNs, its impact on the overall computational cost is minimal. Finally, the TCN-AE module’s computational cost does not increase exponentially with the graph size but remains relatively stable, meaning it will not become a computational bottleneck during training.

In terms of space complexity, our framework does not show significant improvement in memory usage when processing large-scale parking networks. In fact, it even leads to an average increase of 22\% in memory consumption. This increase mainly comes from the following two factors: 1. Graph Coarsening, the increased memory consumption in this module is primarily due to the need to store multiple intermediate results. For example, it requires maintaining the mapping between the original and coarsened graphs, storing clustering labels and cluster centers during the clustering process, and storing intermediate results such as adjacency matrices and Laplacian matrices for each layer of multi-level graph coarsening. 2. TCN-AE, this module requires storing a large amount of intermediate activation values and hidden states during feature compression and reconstruction. Particularly when handling long time-series data, AE needs to store the hidden states and corresponding activation values for each time step, which results in significant memory consumption.

This experiment indicates that while our framework effectively reduces computational costs, it inevitably increases memory consumption due to the introduction of techniques such as graph coarsening. However, we consider this additional memory overhead acceptable, as it is necessary for optimizing computational performance and improving the model's prediction accuracy. Current parking prediction tasks generally have low requirements for memory storage, especially when handling large-scale spatiotemporal data, where memory usage typically does not become a bottleneck. In contrast, time efficiency has a more significant impact on parking prediction performance, as increased computation time often significantly affects model deployment and real-time prediction capabilities. Therefore, we adopted a "trading space for time" optimization strategy, leveraging moderate memory costs to achieve more efficient computational performance. This approach enables the framework to effectively address large-scale urban parking prediction tasks, enhancing training efficiency and prediction accuracy while maintaining high precision.

\section{Conclusion}
In this study, we propose an innovative large-scale urban parking prediction framework based on real-time service capabilities. Specifically, we first introduce a GAT mechanism based on real-time service capabilities, constructing a dynamic parking graph that reflects real-time parking behavior preferences. This aids spatiotemporal GCNs models in more accurately capturing the interactions and trends of change between parking lots. Subsequently, we designed a large-scale urban parking prediction framework based on coarsening and TCN-AE, aimed at effectively reducing the training time required for models to process complex parking data. Finally, we conducted experiments on a real dataset from Shenzhen's parking lots to validate the performance of our proposed prediction framework. The experimental results show that our framework significantly improves both accuracy and efficiency compared to traditional parking prediction methods. Furthermore, as the graph scale expands, our framework's performance exhibits even more pronounced advantages, underscoring its immense potential in addressing complex urban parking problems in practical applications.

While our research has achieved certain results, future work still needs to focus on improving and further exploring the following issues:
\begin{itemize}
    \item Cross-domain Expansion: This study currently relies on the parking dataset from Shenzhen, which is designed for the specific urban layout and traffic patterns of the city, making its applicability limited. To extend the model to other complex transportation networks (such as high-speed rail, airports, etc.), we plan to design adaptive hub node identification algorithms based on the characteristics of different domains and validate the model's generalization ability with cross-domain datasets. This will allow the framework to better adapt to various transportation systems, improving both prediction accuracy and practical application value.
    \item Memory Overhead Optimization: Currently, our approach employs a space-for-time optimization strategy, which effectively improves computational efficiency during training. However, memory consumption remains a bottleneck that needs further optimization in real-world deployment. Therefore, in future work, we plan to consider pruning the time-convolutional autoencoder by removing unnecessary layers and redundant parameters. At the same time, we will incorporate sparse regularization techniques to further compress feature representations, forcing the model to activate only the most important features, thus reducing storage requirements.
    \item Global Information Aggregation: The PRGAT algorithm in this study mainly relies on feature aggregation from 2nd-order neighbors, neglecting potential connections between distant nodes. For complex transportation networks, similar functional areas in different geographical locations often exhibit similar traffic patterns at different times. Understanding these temporal and spatial pattern similarities will help improve the model's overall understanding of traffic flow and its prediction accuracy. Therefore, in future work, we plan to introduce fractal theory to explore node behavior patterns across multiple time scales, identifying potential spatiotemporal relationships between distant nodes, thereby enhancing the model’s ability to capture spatiotemporal dependencies and improving overall prediction performance.
\end{itemize}

\section*{Acknowledgments}
% This should be a simple paragraph before the References to thank those individuals and institutions who have supported your work on this article.
This work was supported in part by Guangdong S\&T Programme (Grant No. 2024B0101020004) and Major Program of Science and Technology of Shenzhen(Grant No.KJZD20231023100304010, KJZD20231023100509018).

% \bibliography{ref}

\begin{thebibliography}{99}
% % % \bibliography{ref}


\bibitem{1}H. Xin, “China registers 415 million motor vehicles, 500 million drivers,” Website, 2024, https://english.www.gov.cn/archive/statistics/202212/08/content WS6391cafcc6d0a757729e41bc.html.
\bibitem{2}W. Zou, Y. Sun, Y. Zhou, Q. Lu, Y. Nie, T. Sun, and L. Peng, “Limited sensing and deep data mining: A new exploration of developing city-wide parking guidance systems,” IEEE Intelligent Transportation Systems Magazine, vol. 14, no. 1, pp. 198–215, 2020.
\bibitem{3}R. Ke, Y. Zhuang, Z. Pu, and Y. Wang, “A smart, efficient, and reliable parking surveillance system with edge artificial intelligence on iot devices,” IEEE Transactions on Intelligent Transportation Systems, vol. 22, no. 8, pp. 4962–4974, 2020.
\bibitem{4}Research and M. ltd, China Smart Parking Industry Report, 2022. Research In China, 2022.
\bibitem{5}S. R. Rizvi, S. Zehra, and S. Olariu, “Aspire: An agent-oriented smart parking recommendation system for smart cities,” IEEE Intelligent Transportation Systems Magazine, vol. 11, no. 4, pp. 48–61, 2018.
\bibitem{6}A. O. Kotb, Y.-c. Shen, and Y. Huang, “Smart parking guidance, monitoring and reservations: a review,” IEEE Intelligent Transportation Systems Magazine, vol. 9, no. 2, pp. 6–16, 2017.
\bibitem{7}K. Yang, X. Tang, Y. Qin, Y. Huang, H. Wang, and H. Pu, “Comparative study of trajectory tracking control for automated vehicles via model predictive control and robust h-infinity state feedback control,” Chinese Journal of Mechanical Engineering, vol. 34, pp. 1-14, 2021.
\bibitem{8}X. Tang, Y. Yang, T. Liu, X. Lin, K. Yang, and S. Li, “Path planning and tracking control for parking via soft actor-critic under non-ideal scenarios,” IEEE/CAA Journal of Automatica Sinica, 2023.
\bibitem{9}Y. Huang, Y. Dong, Y. Tang, and L. Li, “Leverage multi-source traffic demand data fusion with transformer model for urban parking prediction,” arXiv preprint arXiv:2405.01055, 2024.
\bibitem{10}D. Anand, A. Singh, K. Alsubhi, N. Goyal, A. Abdrabou, A. Vidyarthi, and J. J. Rodrigues, “A smart cloud and iovt-based kernel adaptive filtering framework for parking prediction,” IEEE Transactions on Intelligent Transportation Systems, vol. 24, no. 3, pp. 2737–2745, 2022.
\bibitem{11}Y. Ku, C. Guo, K. Zhang, Y. Cui, H. Shu, Y. Yang, and L. Peng, “Toward directed spatiotemporal graph: A new idea for heterogeneous traffic prediction,” IEEE Intelligent Transportation Systems Magazine, vol. 16, no. 2, pp. 70–87, 2024.
\bibitem{12}J. Li, H. Qu, and L. You, “An integrated approach for the near real-time parking occupancy prediction,” IEEE Transactions on Intelligent Transportation Systems, vol. 24, no. 4, pp. 3769–3778, 2022.
\bibitem{13}C. Shuai, X. Zhang, Y. Wang, M. He, F. Yang, and G. Xu, “Online car-hailing origin-destination forecast based on a temporal graph convolutional network,” IEEE Intelligent Transportation Systems Magazine, 2023.
\bibitem{14}B. Song, H. Zhou, X. Huang, and W. Ma, “Collaborative parking vacancy prediction for cities with partial sensors missing,” IEEE Transactions on Intelligent Transportation Systems, 2024.
\bibitem{15}F. J. Enríquez, J.-M. Mejía-Muñoz, G. Bravo, and O. Cruz-Mejıa, “Smart parking: Enhancing urban mobility with fog computing and machine learning-based parking occupancy prediction,” Applied System Innovation, vol. 7, no. 3, p. 52, 2024.
\bibitem{16}X. Zhou, Y. Zhang, Z. Li, X. Wang, J. Zhao, and Z. Zhang, “Large-scale cellular traffic prediction based on graph convolutional networks with transfer learning,” Neural Computing and Applications, pp. 1–11, 2022.
\bibitem{17}M. Fahrbach, G. Goranci, R. Peng, S. Sachdeva, and C. Wang, “Faster graph embeddings via coarsening,” in international conference on machine learning. PMLR, 2020, pp. 2953–2963.
\bibitem{18}C. Cai, D. Wang, and Y. Wang, “Graph coarsening with neural networks,” arXiv preprint arXiv:2102.01350, 2021.
\bibitem{19}A. Loukas, “Graph reduction with spectral and cut guarantees,” Journal of Machine Learning Research, vol. 20, no. 116, pp. 1–42, 2019.
\bibitem{20}M. Kumar, A. Sharma, and S. Kumar, “A unified framework for optimization-based graph coarsening,” Journal of Machine Learning Research, vol. 24, no. 118, pp. 1–50, 2023.
\bibitem{21}Y. Jin, A. Loukas, and J. JaJa, “Graph coarsening with preserved spectral properties,” in International Conference on Artificial Intelligence and Statistics. PMLR, 2020, pp. 4452–4462.
\bibitem{22}Y. Chen, T. Shu, X. Zhou, X. Zheng, A. Kawai, K. Fueda, Z. Yan, W. Liang, I. Kevin, and K. Wang, “Graph attention network with spatial-temporal clustering for traffic flow forecasting in intelligent transportation system,” IEEE Transactions on Intelligent Transportation
Systems, 2022.
\bibitem{23}Z. Lu, W. Lv, Z. Xie, B. Du, G. Xiong, L. Sun, and H. Wang, “Graph sequence neural network with an attention mechanism for traffic speed prediction,” ACM Transactions on Intelligent Systems and Technology (TIST), vol. 13, no. 2, pp. 1-24, 2022.
\bibitem{24}Q. Lu, Z. Tang, Y. Nie, and L. Peng, “Parkingrank-d: A spatial-temporal ranking model of urban parking lots in city-wide parking guidance system,” in 2019 IEEE Intelligent Transportation Systems Conference (iTSC). IEEE, 2019, pp. 388-393.
\bibitem{25}S. Dong, M. Chen, L. Peng, and H. Li, “Parking rank: A novel method of parking lots sorting and recommendation based on public information,” in 2018 IEEE International Conference on Industrial Technology (ICIT). IEEE, 2018, pp. 1381–1386.
\bibitem{26}C.-J. Huang, H. Ma, Q. Yin, J.-F. Tang, D. Dong, C. Chen, G.-Y. Xiang, C.-F. Li, and G.-C. Guo, “Realization of a quantum autoencoder for lossless compression of quantum data,” Physical Review A, vol. 102, no. 3, p. 032412, 2020.
\bibitem{27}M. Liu, T. Zhu, J. Ye, Q. Meng, L. Sun, and B. Du, “spatiotemporal autoencoder for traffic flow prediction,” IEEE Transactions on Intelligent Transportation Systems, 2023.
\bibitem{28}F. Fainstein, J. Catoni, C. P. Elemans, and G. B. Mindlin, “The reconstruction of flows from spatiotemporal data by autoencoders,” Chaos, Solitons \& Fractals, vol. 176, p. 114115, 2023.
\bibitem{29}G. Guo, W. Yuan, J. Liu, Y. Lv, and W. Liu, “Traffic forecasting via dilated temporal convolution with peak-sensitive loss,” IEEE Intelligent Transportation Systems Magazine, vol. 15, no. 1, pp. 48–57, 2021.
\bibitem{30}W. Du, B. Li, J. Chen, Y. Lv, and Y. Li, “A spatiotemporal hybrid model for airspace complexity prediction,” IEEE Intelligent Transportation Systems Magazine, vol. 15, no. 2, pp. 217–224, 2022.
\bibitem{31}Z. Yao, S. Xia, Y. Li, G. Wu, and L. Zuo, “Transfer learning with spatial–temporal graph convolutional network for traffic prediction,” IEEE Transactions on Intelligent Transportation Systems, 2023.
\bibitem{32}L. Zhao, Y. Song, C. Zhang, Y. Liu, P. Wang, T. Lin, M. Deng, and H. Li, “T-gcn: A temporal graph convolutional network for traffic prediction,” IEEE transactions on intelligent transportation systems, vol. 21, no. 9, pp. 3848–3858, 2019.
\bibitem{33}Y. Sun, X. Jiang, Y. Hu, F. Duan, K. Guo, B. Wang, J. Gao, and B. Yin, “Dual dynamic spatial-temporal graph convolution network for traffic prediction,” IEEE Transactions on Intelligent Transportation Systems, vol. 23, no. 12, pp. 23 680-23 693, 2022.
\bibitem{34}D. Li and J. Lasenby, “Spatiotemporal attention-based graph convolution network for segment-level traffic prediction,” IEEE Transactions on Intelligent Transportation Systems, vol. 23, no. 7, pp. 8337-8345, 2021.
\bibitem{35}R. Wickman, X. Zhang, and W. Li, “A generic graph sparsification framework using deep reinforcement learning,” arXiv preprint arXiv:2112.01565, 2021.
\bibitem{36}V. Sadhanala, Y.-X. Wang, and R. Tibshirani, “Graph sparsification approaches for laplacian smoothing,” in Artificial Intelligence and Statistics. PMLR, 2016, pp. 1250-1259.
\bibitem{37}H.-Y. Wu and Y.-L. Chen, “Graph sparsification with generative adversarial network,” in 2020 IEEE International Conference on Data Mining (ICDM). IEEE, 2020, pp. 1328-1333.
\bibitem{38}J. Chen, Y. Saad, and Z. Zhang, “Graph coarsening: from scientific computing to machine learning,” SeMA Journal, vol. 79, no. 1, pp. 187-223, 2022.
\bibitem{39}I. S. Dhillon, Y. Guan, and B. Kulis, “Weighted graph cuts without eigenvectors a multilevel approach,” IEEE transactions on pattern analysis and machine intelligence, vol. 29, no. 11, pp. 1944-1957, 2007.
\bibitem{40}D. Rhouma and L. Ben Romdhane, “An efficient multilevel scheme for coarsening large scale social networks,” Applied Intelligence, vol. 48, pp. 3557-3576, 2018.
\bibitem{41}Q. Wang, H. Jiang, M. Qiu, Y. Liu, and D. Ye, “Tgae: Temporal graph autoencoder for travel forecasting,” IEEE Transactions on Intelligent Transportation Systems, 2022.
\bibitem{42}R. Mo, Y. Pei, N. V. Venkatarayalu, P. N. Joseph, A. B. Premkumar, S. Sun, and S. K. K. Foo, “Unsupervised tcn-ae-based outlier detection for time series with seasonality and trend for cellular networks,” IEEE Transactions on Wireless Communications, vol. 22, no. 5, pp. 3114-3127, 2022.
\bibitem{43}P. Veličković, G. Cucurull, A. Casanova, A. Romero, P. Lio, and Y. Ben-gio, “Graph attention networks,” arXiv preprint arXiv:1710.10903, 2017.
\bibitem{44}B. Yu, H. Yin, and Z. Zhu, “spatiotemporal graph convolutional networks: A deep learning framework for traffic forecasting,” arXiv preprint arXiv:1709.04875, 2017.
\bibitem{45}C. Song, Y. Lin, S. Guo, and H. Wan, “Spatial-temporal synchronous graph convolutional networks: A new framework for spatial-temporal network data forecasting,” in Proceedings of the AAAI conference on artificial intelligence, vol. 34, no. 01, 2020, pp. 914-921.
\bibitem{46}Y. Wang, Y. Ku, Q. Liu, Y. Yang, and L. Peng, “Large-scale parking data prediction: From a graph coarsening perspective,” in 2023 IEEE 26th International Conference on Intelligent Transportation Systems (ITSC). IEEE, 2023, pp. 1410–1415.
\end{thebibliography}

\bibliographystyle{IEEEtran}

% \newpage

 
\vspace{11pt}

\begin{IEEEbiography}[{\includegraphics[width=1in,height=1.25in,clip,keepaspectratio]{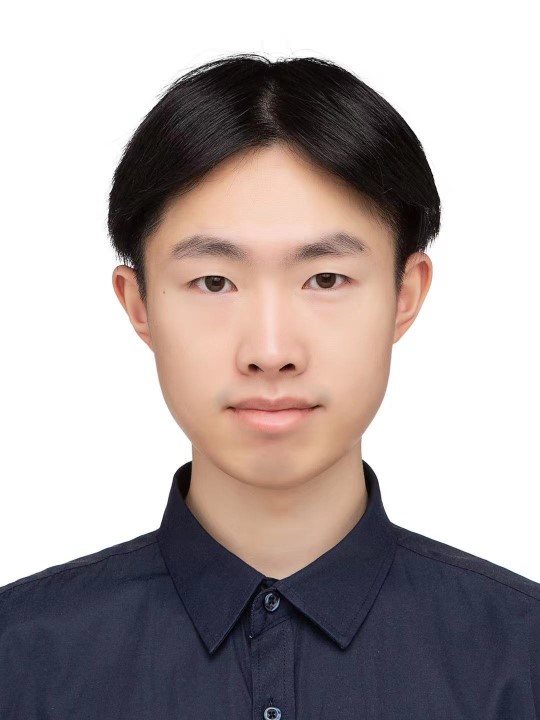}}]{Yixuan Wang}
(yx.wang5@siat.ac.cn) received his B.E. degree in net engineering from Nanjing Agricultural University, Nanjing, China in 2021. He is currently pursuing the M.S. degree in computer science at Shenzhen Institutes of Advanced Technology, Chinese Academy of Sciences, Shenzhen, 518055, as well as University of Chinese Academy of Sciences, Beijing, 100049. His research interests include intelligent transportation systems, spatiotemporal data mining.\end{IEEEbiography}

\begin{IEEEbiography}
[{\includegraphics[width=1in,height=1.25in,clip,keepaspectratio]{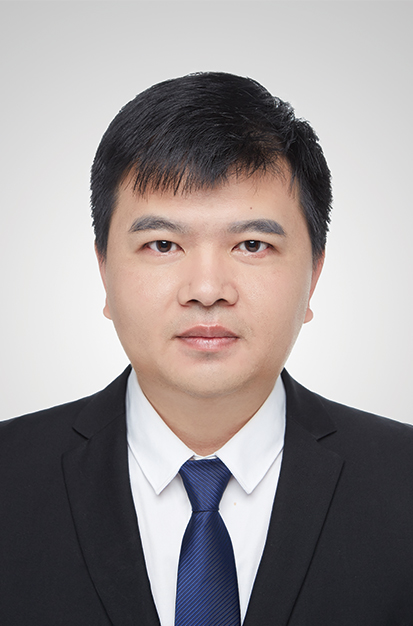}}]{Zhenwu Chen}
(czw@sutpc.com) graduated from the School of Transportation and Engineering of Tongji University of China and received his master's degree in transportation planning and management in 2009. He currently works at Shenzhen Urban Transport Planning Center Co.,Ltd , as the Director of Science and Technology Innovation Center. He is mainly engaged in traffic modeling and simulation, traffic big data, and V2I intelligent traffic control.\end{IEEEbiography}

\begin{IEEEbiography}[{\includegraphics[width=1in,height=1.25in,clip,keepaspectratio]{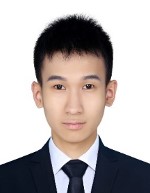}}]{Kangshuai Zhang}
(ks.zhang@siat.ac.cn) received his M.S. degree in mechanical engineering from South China University of Technology, Guangzhou, China in 2021. He is currently a research assistant with Shenzhen Institutes of Advanced Technology, Chinese Academy of Sciences, Shenzhen,518055. His research interests include IoT, ITS, big data mining.\end{IEEEbiography}

\begin{IEEEbiography}[{\includegraphics[width=1in,height=1.25in,clip,keepaspectratio]{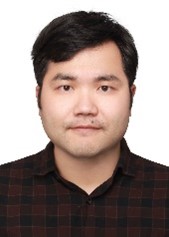}}]{Yunduan Cui}
(yd.cui@siat.ac.cn) earned his Ph.D. degree in computer science from Nara Institute of Science and Technology, Japan.He is currently an associate professor with Shenzhen Institutes of Advanced Technology, Chinese Academy of Sciences, Shenzhen, 518055. His research focuses on the machine learning, intelligent transportation systems and automated driving. He is a Member of IEEE.\end{IEEEbiography}

\begin{IEEEbiography}[{\includegraphics[width=1in,height=1.25in,clip,keepaspectratio]{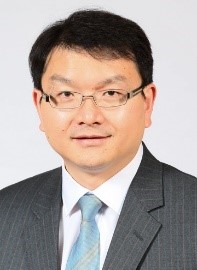}}]{Yang Yang}
(yyiot@hkust-gz.edu.cn) received his PhD degree in Information Engineering from the Chinese University of Hong Kong in 2002. He is currently a professor with the IoT Thrust and the Research Center for Digital World with Intelligent Things (DOIT) at HKUST (Guangzhou), China. His research interests include collaborative intelligence, urban informatics applications. He has published more than 300 papers and filed more than 80 technical patents in these research areas. Yang is a fellow of IEEE.\end{IEEEbiography}

\begin{IEEEbiography}[{\includegraphics[width=1in,height=1.25in,clip,keepaspectratio]{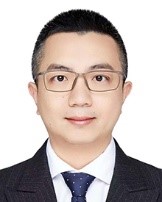}}]{Lei Peng}
(lei.peng@siat.ac.cn) earned his Ph. D degree in computer science from University of Electronic Science and Technology of China. He is currently an associate professor with Shenzhen Institutes of Advanced Technology, Chinese Academy of Sciences, Shenzhen, 518055. His research interests include artificial intelligence, intelligent transportation systems, internet of things and spatiotemporal data mining. He has published more than 60 papers and filed more than 30 technical patents in these research areas. He is a member of IEEE.\end{IEEEbiography}

\vfill

\end{document}